%% file: main.tex
\documentclass[10pt,journal,compsoc]{IEEEtran}
\ifCLASSOPTIONcompsoc
  \usepackage[nocompress]{cite}
\else
  \usepackage{cite}
\fi
\ifCLASSINFOpdf
\else
\fi

\usepackage{enumitem}
\usepackage{amsmath,amsfonts}
\usepackage{amssymb}
\usepackage{algorithm}
\usepackage{algpseudocode}
\usepackage{array}
\usepackage[caption=false,font=normalsize,labelfont=sf,textfont=sf]{subfig}
\usepackage{textcomp}
\usepackage{stfloats}
\usepackage{url}
\usepackage{verbatim}
\usepackage{booktabs}       
\usepackage{graphicx}
\usepackage{multirow}
\usepackage{multicol}
\usepackage{bm}
\usepackage[numbers]{natbib}
\usepackage{colortbl}
\usepackage{xcolor}
\definecolor{maroon}{cmyk}{0,0.87,0.68,0.32}
\definecolor{cvprblue}{rgb}{0.21,0.49,0.74}
\definecolor{algo_comment}{rgb}{0.21,0.4,0.7}
\usepackage[pagebackref,breaklinks,colorlinks,citecolor=cvprblue]{hyperref}
\usepackage{cleveref}

\begin{document}
%
\title{
SVGDreamer++: Advancing Editability and Diversity in Text-Guided SVG Generation
}

%
%

\author{
Ximing~Xing,
Qian~Yu$^\dagger$,~\IEEEmembership{Member,~IEEE,}
Chuang~Wang,
Haitao~Zhou,\\
Jing~Zhang,~\IEEEmembership{Member,~IEEE,}
and~Dong~Xu,~\IEEEmembership{Fellow,~IEEE}
\IEEEcompsocitemizethanks{
\IEEEcompsocthanksitem X. Xing, Q. Yu, C. Wang, H. Zhou, and J. Zhang are with School of Software, Beihang University, Beijing, China (email: ximingxing@buaa.edu.cn, qianyu@buaa.edu.cn, chuangwang@buaa.edu.cn, 18377221@buaa.edu.cn, zhang\_jing@buaa.edu.cn).
\IEEEcompsocthanksitem D. Xu is with Department of Computer Science, The University of Hong Kong, Hong Kong, China (email: dongxu@cs.hku.hk).
\IEEEcompsocthanksitem $^\dagger$ Corresponding author: Qian Yu
}
}

\input{sec/0_abstract}

\maketitle

\input{sec/1_intro}
\input{sec/2_related}
\input{sec/3_method}
\input{sec/4_experiment}
\input{sec/5_conclusion}

\newpage
\bibliographystyle{IEEEtran}
\bibliography{main}


\IEEEdisplaynontitleabstractindextext

%
\IEEEpeerreviewmaketitle

\end{document}

%% file: sec/0_abstract.tex
\IEEEtitleabstractindextext{
\begin{abstract}
Recently, text-guided scalable vector graphics (SVG) synthesis has demonstrated significant potential in domains such as iconography and sketching. However, SVGs generated from existing Text-to-SVG methods often lack editability and exhibit deficiencies in visual quality and diversity. In this paper, we propose a novel text-guided vector graphics synthesis method to address these limitations. To enhance the editability of output SVGs, we introduce a Hierarchical Image VEctorization (HIVE) framework that operates at the semantic object level and supervises the optimization of components within the vector object. This approach facilitates the decoupling of vector graphics into distinct objects and component levels. Our proposed HIVE algorithm, informed by image segmentation priors, not only ensures a more precise representation of vector graphics but also enables fine-grained editing capabilities within vector objects. To improve the diversity of output SVGs, we present a Vectorized Particle-based Score Distillation (VPSD) approach. VPSD addresses over-saturation issues in existing methods and enhances sample diversity. A pre-trained reward model is incorporated to re-weight vector particles, improving aesthetic appeal and enabling faster convergence. Additionally, we design a novel adaptive vector primitives control strategy, which allows for the dynamic adjustment of the number of primitives, thereby enhancing the presentation of graphic details. Extensive experiments validate the effectiveness of the proposed method, demonstrating its superiority over baseline methods in terms of editability, visual quality, and diversity. We also show that our new method supports up to six distinct vector styles, capable of generating high-quality vector assets suitable for stylized vector design and poster design. Code and demo will be released at: \href{http://ximinng.github.io/SVGDreamerV2Project/}{http://ximinng.github.io/SVGDreamerV2Project/}
\end{abstract}
}

%% file: sec/1_intro.tex
\IEEEraisesectionheading{\section{Introduction}\label{sec:introduction}}
\IEEEPARstart{S}{calable} Vector Graphics (SVGs) represent visual concepts using geometric primitives such as B\'ezier curves, polygons, and lines. Due to their inherent nature, SVGs are highly suitable for visual design applications, such as posters and logos. 
Secondly, compared to raster images, vector images can maintain compact file sizes, making them more efficient for storage and transmission purposes. More importantly, vector images offer greater editability, allowing designers to easily select, modify, and compose elements. This attribute is particularly crucial in the design process, as it allows for seamless adjustments and creative exploration.

In recent years, there has been a growing interest in general vector graphics generation. 
Several optimization-based methods have been proposed~\cite{clipdraw_frans_2022,Styleclipdraw_schaldenbr_2022,clipclop_mirowski_2022,Clipasso_vinker_2022,CLIPascene_vinker_2023,CLIPFont_Song_2022,vectorfusion_jain_2023,diffsketcher_xing_2023,svgdreamer_xing_2023,supersvg_hu_2024,NIVeL_thamizharasan_2024,T2VecNeualPath_zhang_2024}, building upon the differentiable rasterizer DiffVG~\cite{diffvg_Li_2020}. These methods, such as CLIPDraw~\cite{clipdraw_frans_2022} and VectorFusion~\cite{vectorfusion_jain_2023}, differ primarily in their approach to supervision.
Some works~\cite{clipdraw_frans_2022,Styleclipdraw_schaldenbr_2022,clipclop_mirowski_2022,CLIPFont_Song_2022,Clipasso_vinker_2022,CLIPascene_vinker_2023} combine the CLIP model~\cite{CLIP_radford_2021} with DiffVG~\cite{diffvg_Li_2020}, using CLIP as a source of supervision. 
More recently, the significant progress achieved by Text-to-Image (T2I) diffusion models~\cite{GLIDE_2022_nichol,ldm_Rombach_2022,DALLE2_2022_ramesh,imagen_2022_saharia,deepfloydif_stability_2023} has inspired the task of Text-to-SVGs. Both VectorFusion~\cite{vectorfusion_jain_2023} and DiffSketcher~\cite{diffsketcher_xing_2023} attempted to utilize T2I diffusion models for supervision. These models make use of the high-quality raster images generated by T2I models as targets to optimize the parameters of vector graphics. Additionally, the priors embedded within T2I models can be distilled and applied in this task. 
Consequently, models that use T2I for supervision generally perform better than those using the CLIP model.

Despite their impressive performance, existing T2I-based methods have certain limitations. Firstly, the vector images generated by these methods lack editability. Unlike the conventional approach to creating vector graphics, where individual elements are added one by one, T2I-based methods do not distinguish between different components during synthesis. As a result, the objects become entangled, making it challenging to edit or modify a single object independently, let alone make changes to local details.
Secondly, there is still a large room for improvement in visual quality and diversity of the results generated by these methods. 
Both VectorFusion~\cite{vectorfusion_jain_2023} and DiffSketcher~\cite{diffsketcher_xing_2023} extended the Score Distillation Sampling (SDS)~\cite{dreamfusion_poole_2023} to distill priors from the T2I models.
However, it has been observed that SDS can lead to issues such as color over-saturation and over-smoothing, resulting in a lack of fine details in the generated vector images.
Besides, SDS optimizes a set of control points in the vector graphic space to obtain the average state of the vector graphic corresponding to the text prompt in a mode-seeking manner~\cite{dreamfusion_poole_2023}. 
This leads to a lack of diversity and details in the SDS-based approach~\cite{vectorfusion_jain_2023,diffsketcher_xing_2023}, along with absent text prompt objects.

To address the aforementioned issues, we present a new approach called \textit{SVGDreamer} for text-guided vector graphics generation. Our primary objective is to produce vector graphics of superior quality that offer enhanced editability, visual appeal, and diversity.
To ensure editability, we propose a \textbf{S}emantic-driven \textbf{I}mage \textbf{VE}ctorization (SIVE) process. This approach incorporates an innovative attention-based primitive control strategy, which facilitates the decomposition of the synthesis process into foreground objects and background.
To initialize the control points for each foreground object and background, we leverage cross-attention maps queried by text tokens.
Furthermore, we introduce an attention-mask loss function, which optimizes the graphic elements hierarchically. The proposed SIVE process ensures the separation and editability of object-level elements, promoting effective control and manipulation of the resulting vector graphics.

To improve the visual quality and diversity of the generated vector graphics, we introduce Vectorized Particle-based Score Distillation (VPSD) for vector graphics refinement.
Previous works in vector graphics synthesis~\cite{vectorfusion_jain_2023,diffsketcher_xing_2023,wordasimg_Iluz_2023} that utilized SDS often encountered issues like shape over-smoothing, color over-saturation, limited diversity, and slow convergence in synthesized results~\cite{dreamfusion_poole_2023, diffsketcher_xing_2023}.
To address these issues, VPSD models SVGs as distributions of control points and colors, respectively.
VPSD adopts a LoRA~\cite{lora_hu_2022} network to estimate these distributions, aligning vector graphics with the pretrained diffusion model.
Furthermore, to enhance the aesthetic appeal of the generated vector graphics, we integrate Reward Feedback Learning (ReFL) ~\cite{imagereward_xu_2023} to fine-tune the estimation network. 
Through this refinement process, we achieve the final vector graphics with a more human aesthetic evaluation.


Building upon our previous exploration, we introduce an enhanced approach termed \textit{SVGDreamer++}, which offers significant improvements over SVGDreamer, particularly in two key aspects:
\textit{Firstly}, we introduce a \textbf{H}ierarchical \textbf{I}mage \textbf{VE}ctorization (HIVE) strategy to enhance the visual quality and editability of synthesized SVGs, particularly in fine details.
While SIVE focuses on object-level decomposition of the output SVGs using attention maps for guidance, HIVE employs image segmentation priors to control both object-level and part-level elements, thereby producing more accurate boundaries in synthesized SVGs.
Specifically, HIVE synergizes a diffusion model with the segmentation model SAM~\cite{SegmentAnything_kirillov_2023}, leveraging attention priors of the diffusion model to condition SAM for more precise masks.
\textit{Secondly}, we propose \textit{Adaptive Vector Primitive Control}, a new algorithm that dynamically adjusts the number of vector primitives during the optimization phase. 
The number of vector paths is crucial for the visual quality of generated SVGs.
However, setting an optimal count is challenging: too few paths may degrade geometric features, 
while too many can slow the optimization process.
For the first time, we investigate the adaptive adjustment of the number of primitives based on the content of the image. 
This capability allows our SVGDreamer++ to achieve superior visual quality. 
By synergizing HIVE, Adaptive Vector Primitive Control, and VPSD within the SVGDreamer++ framework, we enable the creation of high-quality, editable, and diverse vector graphics.

Extensive experiments are conducted to validate the effectiveness of SVGDreamer++, demonstrating its superiority over baseline methods in terms of editability, visual quality, and diversity.
SVGDreamer++ supports up to six distinct vector styles, and our experiments indicate that it can generate high-quality vector assets suitable for stylized vector design.
Furthermore, we demonstrate the applicability of our approach in vector design, including icon creation and poster design.

\begin{figure*}[!t]
\centering
\includegraphics[width=1.0\linewidth]{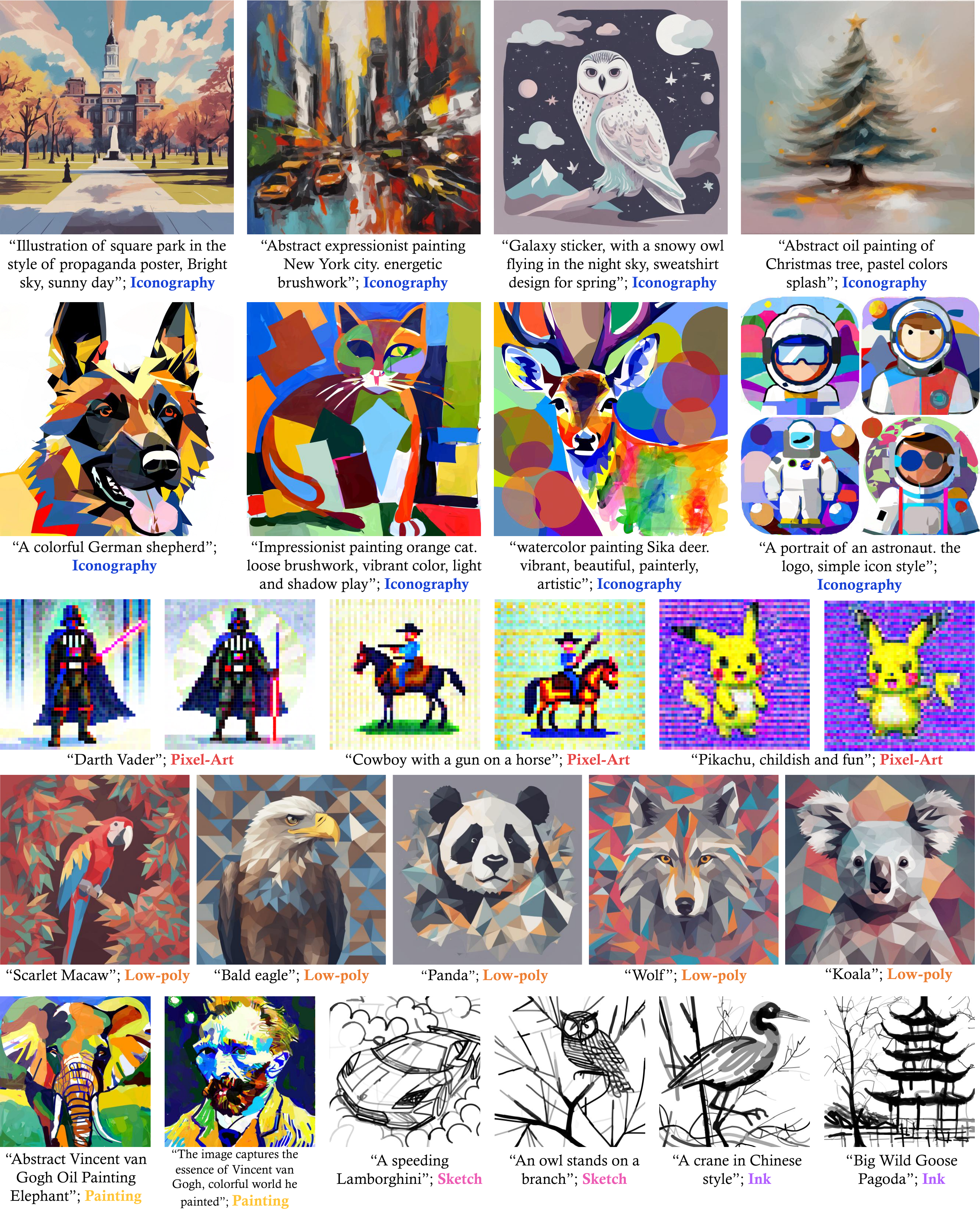}
\vspace{-1em}
\caption{
\textbf{SVGs produced by SVGDremaer++.} Given a text prompt, SVGDreamer++ can generate a variety of vector graphics. SVGDreamer++ is a versatile tool that can work with various vector styles without being limited to a specific prompt suffix. We utilize various colored suffixes to indicate different styles. The style is governed by vector primitives.
} \label{fig:gallery}
\end{figure*}

Parts of the results in this paper were originally published in its conference version~\cite{svgdreamer_xing_2023}. 
Furthermore, this paper extends our earlier work in several important aspects:
\begin{itemize}[leftmargin=*]
\item An enhanced SVGDreamer++ approach is introduced, specifically tailored for text-to-SVG generation. This novel approach is capable of producing vector graphics with better visual quality and higher editability (Figure~\ref{fig:gallery}, \ref{fig:editable}).
\item In SVGDreamer++, we introduce a Hierarchical Image VEctorization technique (HIVE, Section~\ref{sec:hive}) that facilitates composition decoupling and localized editing of vector objects.
We conduct new experiments to investigate the effectiveness of HIVE (Figures~\ref{fig:editable}, \ref{fig:abl_live_vs_sive_vs_five}).
\item We propose a plug-and-play \textit{Adaptive Vector Primitives Control} algorithm for SVGDreamer++ (Section~\ref{sec:adaptive_path_control}, Algorithm~\ref{algo:adaptive_vec_control}). 
This method optimizes the performance of SVGDreamer++ by addressing missing geometrical features in vector graphs, 
thus achieving better visual quality (Figures~\ref{fig:path_control_pipe}, \ref{fig:hive_grad_loss_trace}).
Substantial experimental results support the effectiveness of this approach (Figure~\ref{fig:abl_path_control}, \ref{fig:path_control_process}).
\item We conduct comprehensive experiments to demonstrate the effectiveness of our newly proposed components. We also provide more qualitative and quantitative results (Table~\ref{tab:quantitative}, Figure~\ref{fig:compare_methods}, \ref{fig:diverse_results}, \ref{fig:vector_assets}) to show the superiority of our proposed SVGDreamer++ over baseline methods.
\end{itemize}

%% file: sec/2_related.tex
\section{Related Work}
\label{sec:related_work}
\subsection{Vector Graphics Generation}
\noindent Scalable Vector Graphics (SVGs) provide a declarative format for visual concepts articulated through primitives. SVGs are extensively utilized in the design domain owing to their manipulable geometric composition, resolution independence, and compact file size.
One approach to generating SVG content entails training a neural network to generate predefined SVG commands and attributes~\cite{sketchrnn_David_2018,SVGVAE_Lopes_2019,deepsvg_carlier_2020,im2vec_reddy_2021,deepvecfont_wang_2021,iconshop_wu_2023,strokenuwa_tang_2024}. 
Neural networks designed for learning SVG representations typically include architectures such as RNNs~\cite{sketchrnn_David_2018,im2vec_reddy_2021}, VAEs~\cite{SVGVAE_Lopes_2019}, and Transformers~\cite{deepsvg_carlier_2020,deepvecfont_wang_2021,iconshop_wu_2023,strokenuwa_tang_2024}. The training of these networks is heavily dependent on datasets in vector form. However, the limited availability of large-scale vector datasets significantly constrains their generalization capability and their ability to synthesize intricate vector graphics.
To date, the domain of vector graphics has not benefited from datasets of a scale comparable to ImageNet~\cite{ImageNet}. The existing datasets are predominantly focused on specific, narrow areas, such as monochromatic (black-and-white) vector icons~\cite{FIGR_clouatre_2019,deepsvg_carlier_2020}, emojis~\cite{NotoEmoji_google_2022} and fonts~\cite{SVGVAE_Lopes_2019}.
Instead of directly learning an SVG generation network, an alternative method of vector synthesis is to optimize towards a matching image during evaluation time.

Li \textit{et al.}~\cite{diffvg_Li_2020} introduce a differentiable rasterizer that bridges the vector graphics and raster image domains. 
While image generation methods that traditionally operate over vector graphics require a vector-based dataset, recent works has demonstrated the use of differentiable rasterizer to overcome this limitation~\cite{ClipGen_Shen_2022, evolution_tian_2022, Styleclipdraw_schaldenbr_2022, LIVE_Ma_2022, marvel_su_2023, CLIPVG_song_2023, diffsketcher_xing_2023,supersvg_hu_2024,NIVeL_thamizharasan_2024,T2VecNeualPath_zhang_2024}. 
This approach for SVG generation involves directly optimizing the geometric and color parameters of SVG paths using the guidance of a pretrained vision-language model.
Recent advances in visual text embedding contrastive language-image pre-training model (CLIP)~\cite{CLIP_radford_2021} have enabled a number of successful methods for synthesizing sketches, such as CLIPDraw~\cite{clipdraw_frans_2022}, CLIP-CLOP~\cite{clipclop_mirowski_2022}, and CLIPasso~\cite{Clipasso_vinker_2022}. 
In contrast to CLIP, the diffusion model demonstrates superior generation abilities and exhibits enhanced image consistency.
VectorFusion~\cite{vectorfusion_jain_2023} and DiffSketcher~\cite{diffsketcher_xing_2023} integrate differentiable rasterizers with text-to-image diffusion models to generate vector graphics, yielding promising results in domains such as iconography, pixel art, and sketching. 
Although the above methods introduce the raster priors of diffusion model into the vector domain beforehand, its editability and graphical quality are insufficient.
Moreover, recent studies~\cite{NIVeL_thamizharasan_2024,T2VecNeualPath_zhang_2024} have combined optimization-based approaches with neural network training to learn vector representations, thereby incorporating geometric constraints into vector graphics.
Our proposed SVGDreamer++ from an alternative approach, bypassing neural network training by utilizing image segmentation priors for enforcing geometric constraints. Concurrently, we introduce a novel plug-and-play vector primitive control method based on optimization.

\subsection{Diffusion Models}
\noindent Denoising diffusion probabilistic models (DDPMs)~\cite{diffusion_models_dickstein_2015,EestGrad_song_2019,ddpm_ho_2020,scorebased_song_2021,ADM_dhariwal_2021,iDDPM_nichol_2021,ddim_song_2021}, particularly those conditioned on text, have shown promising results in text-to-image synthesis. For example, Classifier-Free Guidance (CFG)~\cite{classifierfree_2022_ho} has improved visual quality and is widely used in large-scale text conditional diffusion model frameworks, including GLIDE~\cite{GLIDE_2022_nichol}, Stable Diffusion~\cite{ldm_Rombach_2022}, DALL·E 2~\cite{DALLE2_2022_ramesh}, Imagen~\cite{imagen_2022_saharia} and DeepFloyd IF~\cite{deepfloydif_stability_2023}, SDXL~\cite{sdxl_podell_2024}.
The progress achieved by text-to-image (T2I) diffusion models~\cite{GLIDE_2022_nichol,ldm_Rombach_2022,DALLE2_2022_ramesh,imagen_2022_saharia} also promotes the development of a series of text-guided tasks, such as text-to-3D~\cite{dreamfusion_poole_2023,SJC_Wang_2023} and text-to-video~\cite{VDM_ho_2022,makeAvideo_singer_2023}. 

Recent advances in natural image modeling have sparked significant research interest in utilizing powerful 2D pretrained models to recover 3D object structures~\cite{clipforge_sanghi_2022,dreamfields_jain_2022,GAN2shape_pan_2021,SJC_Wang_2023, Magic3D_Lin_2023,dreamfusion_poole_2023,prolificdreamer_wang_2023}.
Recent efforts such as DreamFusion~\cite{dreamfusion_poole_2023}, Magic3D~\cite{Magic3D_Lin_2023} and Fantasia3D~\cite{Fantasia3D_Chen_2023} explore text-to-3D generation by exploiting a score distillation sampling (SDS) loss derived from a 2D text-to-image diffusion model~\cite{imagen_2022_saharia, ldm_Rombach_2022} instead, showing impressive results.
The development of text-to-SVG~\cite{vectorfusion_jain_2023,diffsketcher_xing_2023} was inspired by this, but the resulting vector graphics have limited quality and exhibit a similar over-smoothness as the reconstructed 3D models.
Wang~\textit{et al.}~\cite{prolificdreamer_wang_2023} extend the modeling of the 3D model as a random variable instead of a constant as in SDS and present variational score distillation to address the over-smoothing issues in text-to-3D generation.

In this work, we extend the T2I model to the domain of vector graphics, facilitating the synthesis of graphics with image-like realism. Furthermore, we illustrate the potential of the proposed method within the realm of vector design.


%% file: sec/3_method.tex
\section{The SVGDreamer Approach}
\label{sec:svgdreamer}
\begin{figure}[t]
\centering
\includegraphics[width=1.0\linewidth]{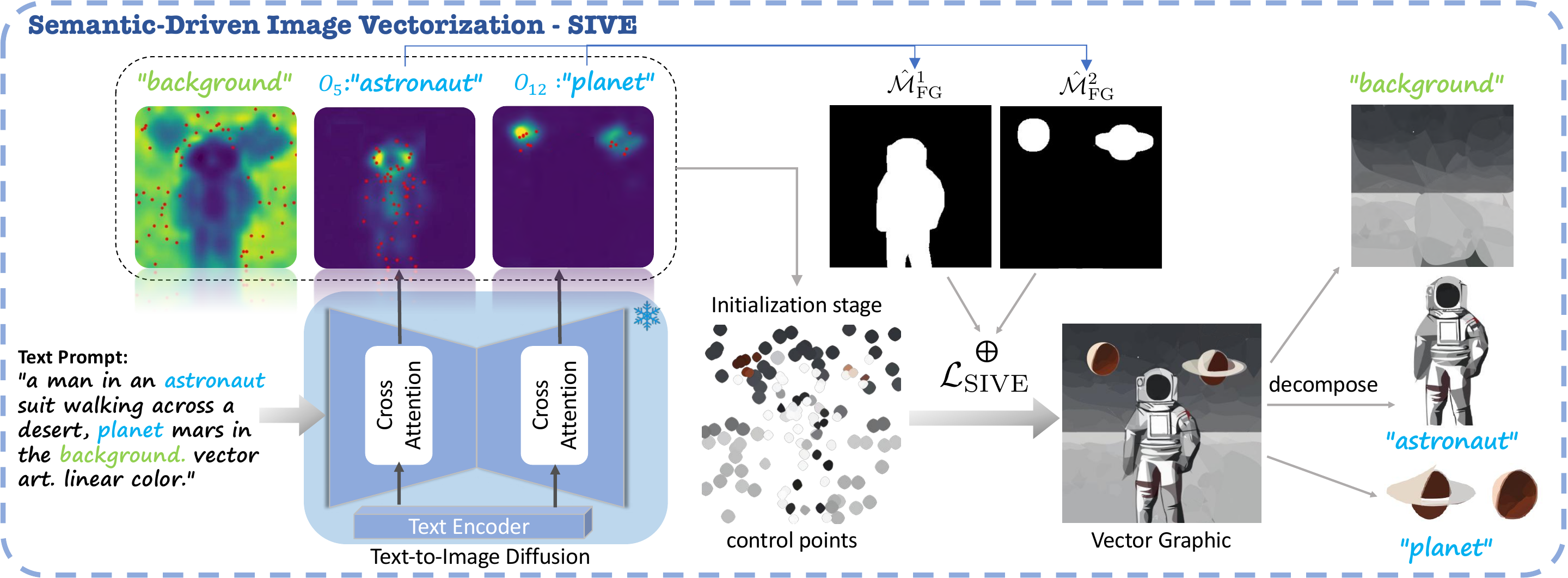}
\vspace{-1.5em}
\caption{
\textbf{The pipeline of SIVE}. 
SIVE comprises two primary modules: primitive initialization and semantic-aware optimization. The primitive initialization module leverages diffusion model attention priors to initially delineate the paths of the corresponding vector objects. Subsequently, an attention-based mask loss function is introduced to facilitate the hierarchical optimization of these vector objects.
} \label{fig:pipe_sive}
\vspace{-1em}
\end{figure}
\noindent In this section, we introduce SVGDreamer, an optimization-based method that creates a variety of vector graphics based on text prompts. 
A vector graphic is defined as a set of paths, $\{P_i\}_{i=1}^n$, and color attributes, $ \{C_i\}_{i=1}^n $. Each path is comprised of $m$ control points, $P_i=\{ p_j \}_{j=1}^m = \{(x_j,y_j) \}_{ j=1 }^m $, and one color attribute, $C_i= \{ r,g,b,a \}_i$.
In this paper, we will optimize the SVG parameters to progressively evolve their initial state into a more refined and accurate graphical representation.
We optimize an SVG by backpropagating gradients of rasterized images to the SVG path parameters,  $\bm{\theta} = \{P_i, C_i\}_{i=1}^n$, utilizing a differentiable renderer~\cite{diffvg_Li_2020} $\mathcal{R}(\bm{\theta})$.

Our approach leverages the pre-trained text-to-image diffusion model prior to guide the differentiable renderer $\mathcal{R}$ and optimize the parametric graphic path $\theta$, resulting in the synthesis of vector graphs that match the description of the text prompt $y$.
Our pipeline consists of two parts: semantic-driven image vectorization (Fig.~\ref{fig:pipe_sive}) and SVG synthesis through VPSD optimization (Fig.~\ref{fig:vpsd_pipeline}).
The first part is \textbf{S}emantic-driven \textbf{I}mage \textbf{VE}ctorization (SIVE), consisting of two stages: primitive initialization and semantic-aware optimization.
We rethink the application of attention mechanisms in synthesizing vector graphics.
We extract the cross-attention maps corresponding to different objects in the diffusion model and apply it to initialize control points and consolidate object vectorization.
This process allows us to decompose the foreground objects from the background.
Consequently, the SIVE process generates vector objects which are independently editable. It separates vector objects by aggregating the curves that form them, which in turn simplifies the combination of vector graphics.

In section~\ref{sec:vpsd}, we propose the \textbf{V}ectorized \textbf{P}article-based \textbf{S}core \textbf{D}istillation (VPSD) to generate diverse high-quality text-matching vector graphics.
VPSD is designed to model the distribution of vector path control points and colors for approximating the vector parameter distribution, thus obtaining vector results of diversity.
\subsection{SIVE: Semantic-driven Image Vectorization}
\label{sec:SIVE}
\noindent Image rasterization is a mature technique in computer graphics, while image vectorization, the reverse path of rasterization, remains a major challenge.
Given an arbitrary input image, LIVE~\cite{LIVE_Ma_2022} recursively learns the visual concepts by adding new optimizable closed Bézier paths and optimizing all these paths.
However, LIVE~\cite{LIVE_Ma_2022} struggles with grasping and distinguishing various subjects within an image, leading to identical paths being superimposed onto different visual subjects. And the LIVE-based method~\cite{LIVE_Ma_2022, vectorfusion_jain_2023}  fails to represent intricate vector graphics consisting of complex paths. 
We propose a semantic-driven image vectorization method to address the aforementioned issue. This method consists of two main stages: primitive initialization and semantic-aware optimization.
In the initialization stage, we allocate distinct control points to different regions corresponding to various visual objects with the guidance of attention maps.
In the optimization stage, we introduce an attention-based mask loss function to hierarchically optimize the vector objects.

\subsubsection{Primitive Initialization}
\label{sec:primitive_init}
Vectorizing visual objects often involves assigning numerous paths, which leads to \textit{object-layer confusion} in LIVE-based methods. 
To address this issue, we suggest organizing vector graphic elements semantically and assigning paths to objects based on their semantics.
We initialize $O$ groups of object-level control points according to the cross-attention map corresponding to different objects in the text prompt. 
And we represent them as the foreground $\mathcal{M}_{\mathrm{FG}}^i$, where $i$ indicates the $i$-th token in the text prompt.
Correspondingly, the rest will be treated as background.
Such design allows us to represent the attention maps of background and foreground as,
\begin{equation}
\mathcal{M}_{\mathrm{BG}} = \mathrm{Inv}(\sum_{i=1}^{O} \mathcal{M}_{\mathrm{FG}}^i); \ 
\mathcal{M}_{\mathrm{FG}}^i = \mathrm{softmax} (Q K^{T}_i) / \sqrt{d}
\label{eq:sive_mask}
\end{equation}
\noindent where $\mathcal{M}_{\mathrm{BG}}$ indicates the attention map of the background.
$\mathrm{Inv}(\cdot)$ indicates the reverse operation of the sum of $\mathcal{M}_{\mathrm{FG}}^i$.
$\mathcal{M}_{\mathrm{FG}}^i$ indicates cross-attention score, where $K_i$ indicates $i$-th token keys from text prompt, $Q$ is pixel queries features, and $d$ is the latent projection dimension of the keys and queries.

Then, inspired by DiffSketcher~\cite{diffsketcher_xing_2023}, we normalize the attention maps using softmax and treat it as a distribution map to sample $m$ positions for the first control point $p_{j=1}$ of each Bézier curve. 
The other control points ($\{p_j\}_{j=2}^m$) are sampled within a small radius (0.05\% of image size) around $p_{j=1}$ to define the initial set of paths.
In the following section, we will explain how to consolidate object semantics during the synthesis of vector graphics using the mask.

\subsubsection{Semantic-aware Optimization}
\label{sec:semantic_aware_optim}
In this stage, we utilize an attention-based mask loss to separately optimize the objects in the foreground and background. 
This ensures that control points remain within their respective regions, aiding in object decomposition. 
Namely, the hierarchy only exists within the designated object and does not get mixed up with other objects.
This strategy fuels the permutations and combinations between objects that form different vector graphics, and enhances the editability of the objects themselves. 

Specifically, we convert the attention map obtained during the initialization stage into masks $\hat{\mathcal{M}} = \{ \{\hat{\mathcal{M}}_{\mathrm{FG}}\}_{o=1}^{O}, \hat{\mathcal{M}}_{\mathrm{BG}} \}$, $O$ foregrounds and one background mask in total. 
This is accomplished by assigning the attention score a value of 1 if it exceeds the predefined threshold, and 0 otherwise. Subsequently, the background mask is generated by inverting the foreground mask, ensuring accurate differentiation between foreground and background regions. Finally, we add mask constraints to the optimization,
\begin{equation}
\mathcal{L}_{\mathrm{SIVE}} = 
\sum_{i}^O \left( \hat{\mathcal{M}}_i \odot I - \hat{\mathcal{M}}_i \odot \bm{x} \right)^2
\label{eq:sive}
\end{equation}
\noindent where $I$ is the target image, $\hat{\mathcal{M}}$ is mask, $\bm{x} = \mathcal{R}(\bm{\theta})$ is the rendering.
  
\subsection{VPSD: Vectorized Particle-based Score Distillation}
\label{sec:vpsd}
\begin{figure}[t]
\centering
\includegraphics[width=1.0\linewidth]{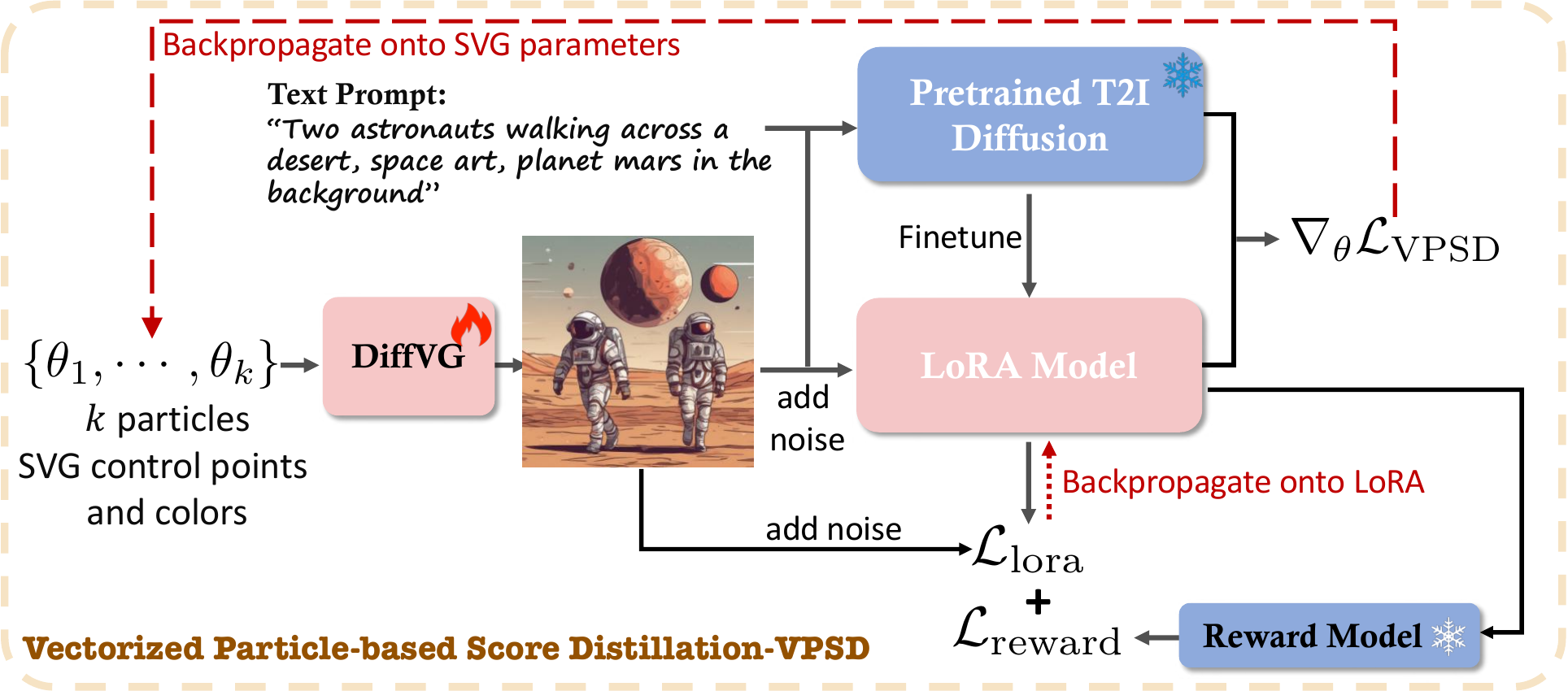}
\vspace{-1.5em}
\caption{
\textbf{The process of Vectorized Particle-based Score Distillation}. VPSD accepts $k$ sets of SVG parameters as input. 
VPSD models SVG as a distribution of vector paths and color parameters, estimating these parameters through the application of the LoRA network. Through the estimation of the SVG parameter distribution, VPSD achieves a greater diversity of outputs compared to VF~\cite{vectorfusion_jain_2023}.  Moreover, to enhance the aesthetic quality of the vector outputs, a pretrained reward model~\cite{imagereward_xu_2023} is employed to optimize the training process of the estimation network.
} \label{fig:vpsd_pipeline}
\vspace{-1em}
\end{figure}
\noindent \textbf{The Diversity of SVG Generation}. While vectorizing a rasterized diffusion sample is lossy, recent techniques~\cite{vectorfusion_jain_2023,diffsketcher_xing_2023} have identified the SDS loss~\cite{dreamfusion_poole_2023} as beneficial for our task of generating vector graphics.
To synthesize a vector image that matches a given text prompt $y$, they directly optimize the parameters $\bm{\theta} = \{P_i, C_i\}_{i=1}^n$ of a differentiable rasterizer $\mathcal{R}(\bm{\theta})$ via SDS loss.
At each iteration, the differentiable rasterizer is used to render a raster image $\bm{x}=\mathcal{R}(\bm{\theta})$, which is then data augmented to obtain $\bm{x}_a$.
Then, the pretrained latent diffusion model (LDM) $\epsilon_\phi$ uses a VAE encoder~\cite{taming_esser_2021} to encode $\bm{x}_a$ into a latent representation $\bm{z} = \mathcal{E}(\bm{x}_a)$, where $\bm{z} \in \mathbb{R}^{(H / f) \times (W / f) \times 4}$ and $f$ is the VAE encoder downsample factor.
Finally, the gradient of SDS is estimated by,
\begin{equation}
\begin{split}
\nabla_{\bm{\theta}} \mathcal{L}_{\mathrm{SDS}} & (\phi, \bm{x} = \mathcal{R}(\bm{\theta})) \triangleq 
\\
& \mathbb{E}_{t,\mathbf{\epsilon},a} 
\left[ 
w(t) (\mathbf{\epsilon}_{\phi} (\bm{z}_t;y,t) - \mathbf{\epsilon}) 
\frac{\partial \mathbf{z}}{\partial \bm{x}_a}
\frac{\partial \mathbf{x}_a}{\partial \theta}
\right]
\end{split}
\end{equation}
\noindent where $w(t)$ is the weighting function. And noised to form $\bm{z}_t = \alpha_t \bm{x}_a + \sigma_t \mathbf{\epsilon}$.

\begin{figure*}[t]
\centering
\includegraphics[width=1.0\linewidth]{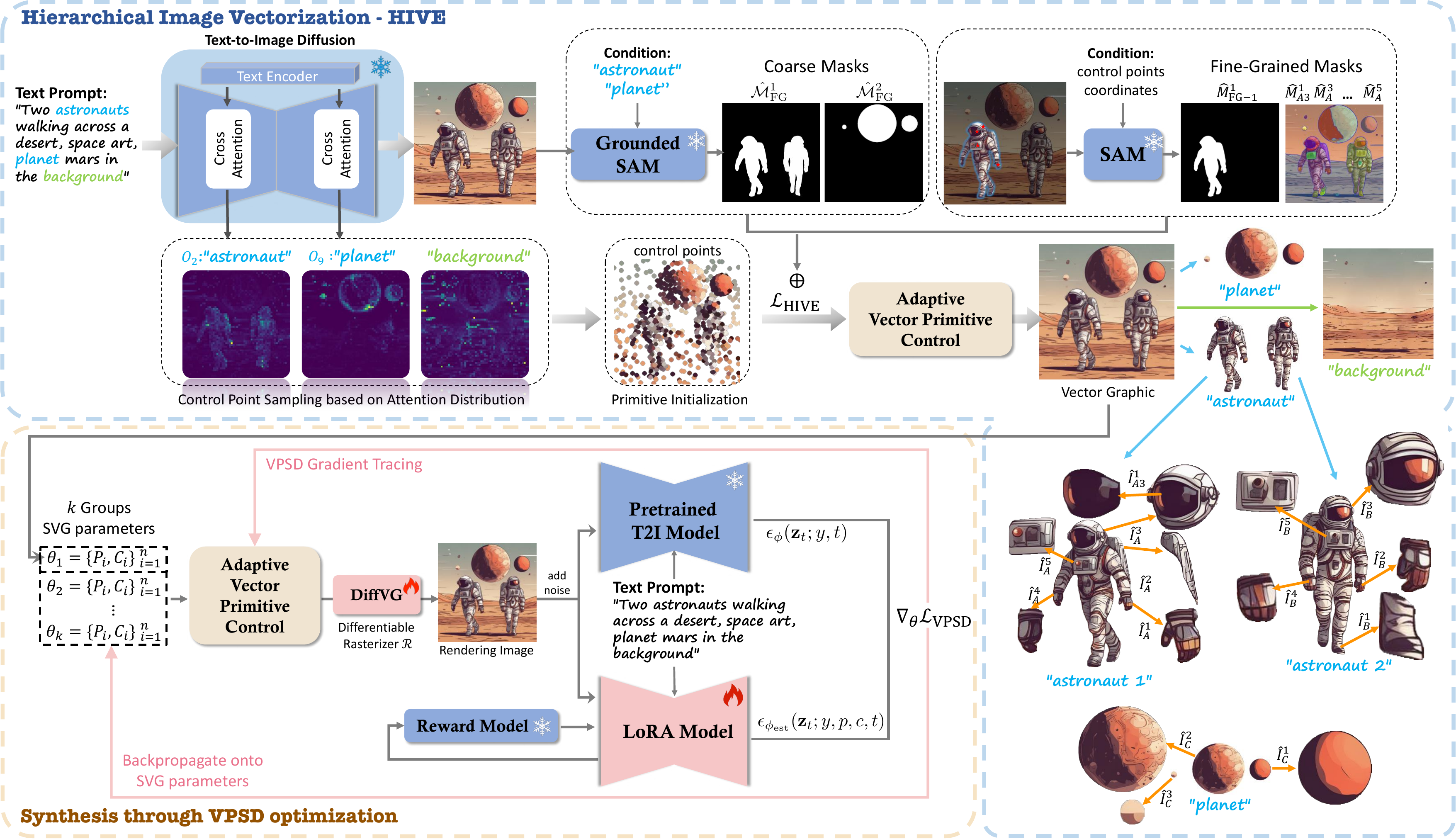}
\vspace{-1.5em}
\caption{
\textbf{Overview of SVGDreamer++.} 
Our method consists of two phases: Hierarchical image vectorization~(Sec.~\ref{sec:hive}) and optimized synthesis of diverse SVGs via VPSD~(Sec.~\ref{sec:vpsd}).
And an additional module, called Adaptive Vector Primitives Control~(Sec.~\ref{sec:adaptive_path_control}), can be plugged into HIVE and VPSD in a plug-and-play way.
In HIVE we introduced two stages of mask generation (as shown in the dotted box). Coarse mask generation guided by prompt words and fine-grained mask generation guided by attention distribution are used to decouple the components of vector graphics.
The result from HIVE can be used as input for further generation of VPSD. We maintain $k$ sets of SVG parameters in VPSD for obtaining diverse results.
In addition, the brown dotted box represents adaptive vector primitive control technology, which dynamically builds vector paths based on gradient graphs to improve the quality of SVG synthesis.
} \label{fig:pipeline}
\vspace{-1em}
\end{figure*}

Unfortunately, SDS-based methods often suffer from issues such as shape over-smoothing, color over-saturation, limited diversity in results, and slow convergence in synthesis results~\cite{dreamfusion_poole_2023, vectorfusion_jain_2023, diffsketcher_xing_2023, wordasimg_Iluz_2023}.
Inspired by the principled variational score distillation framework~\cite{prolificdreamer_wang_2023}, we propose vectorized particle-based score distillation (VPSD) to address the aforementioned issues.
Instead of modeling SVGs as a set of control points and corresponding colors like SDS, we model SVGs as the \textit{distributions} of control points and colors respectively. In principle, given a text prompt $y$, there exists a probabilistic distribution $\mu$ of all possible vector shapes representations.
Under a vector representation parameterized by $\bm{\theta}$, such a distribution can be modeled as a probabilistic density $\mu(\bm{\theta}|y)$.
Compared with SDS that optimizes for the single $\bm{\theta}$, VPSD optimizes for the whole distribution $\mu$, from which we can sample $\theta$.
Motivated by previous particle-based variational inference methods, we maintain $k$ groups of vector parameters $\{ \bm{\theta} \}_{i=1}^k$ as particles to estimate the distribution $\mu$, and $\bm{\theta}_i$ will be sampled from the optimal distribution $\mu^{\ast}$ if the optimization converges. 
This optimization can be realized through two score functions: one that approximates the optimal distribution with a noisy real image, and one that represents the current distribution with a noisy rendered image.
The score function of noisy real images can be approximated by the pretrained diffusion model~\cite{ldm_Rombach_2022} $\mathbf{\epsilon}_{\phi} (\bm{z}_t;y,t)$.
The score function of noisy rendered images is estimated by another noise prediction network $\mathbf{\epsilon}_{\phi_\mathrm{est}}(\bm{z}_t;y,p,c,t)$, which is trained on the rendered images by $\{ \bm{\theta} \}_{i=1}^k$.
The gradient of VPSD can be formed as,
\begin{equation}
\begin{split}
& \nabla_{\bm{\theta}} \mathcal{L}_{\mathrm{VPSD}} (\phi, \phi_\mathrm{est} , \bm{x} = \mathcal{R}(\bm{\theta})) \triangleq 
\\
& \mathbb{E}_{t,\epsilon,p,c} 
\left[ 
w(t) (
\mathbf{\epsilon}_{\phi} (\bm{z}_t;y,t) - 
\mathbf{\epsilon}_{\phi_\mathrm{est}}(\mathbf{z}_t;y,p,c,t) 
)
\frac{\partial \bm{z}}{\partial \bm{\theta}}
\right]
\end{split}
\end{equation}
\noindent where $p$ and $c$ in $\mathbf{\epsilon}_{\phi_\mathrm{est}}$ indicate control point variables and color variables, the weighting function $w(t)$ is a hyper-parameter. And $t \sim \mathcal{U}(0.05, 0.95)$.

In practice, as suggested by~\cite{prolificdreamer_wang_2023}, we parameterize $\mathbf{\epsilon}_{\phi}$ using a LoRA (Low-rank adaptation~\cite{lora_hu_2022}) of the pretrained diffusion model.
The rendered image not only serves to calculate the VPSD gradient but also gets updated by LoRA,
\begin{equation}
\mathcal{L}_{\mathrm{lora}} = 
\mathbb{E}_{t,\epsilon,p,c} 
\left\| \mathbf{\epsilon}_{\phi_\mathrm{est}}(\bm{z}_t;y,p,c,t) - \epsilon \right\|_{2}^{2}
\end{equation}
\noindent where $\epsilon$ is the Gaussian noise. Only the parameters of the LoRA model will be updated, while the parameters of other diffusion models will remain unchanged to minimize computational complexity.

\noindent\textbf{The Aesthetics of SVG Generation}. In~\cite{prolificdreamer_wang_2023}, only randomly selected particles update the LoRA network in each iteration. 
However, this approach neglects the learning progression of vector particles, which are used to represent the optimal SVG distributions. Furthermore, these networks typically require numerous iterations to approximate the theoretical optimal distribution, resulting in slow convergence. 
In VPSD, we introduce a Reward Feedback Learning method, as Fig.~\ref{fig:vpsd_pipeline} illustrates. This method leverages a pre-trained reward model~\cite{imagereward_xu_2023} to assign reward scores to samples collected from LoRA model. Then LoRA model subsequently updates from these reweighted samples,
\begin{equation}
\mathcal{L}_{\mathrm{reward}} = \mathbb{E}_{y} \left[
\mathbf{\psi}( r( y, g_{\phi_{\mathrm{est}}}(y) ) ) 
\right]
\end{equation}
\noindent where $g_{\phi_{\mathrm{est}}}(y)$ denotes the generated image of $\mu$ model with parameters $\phi_{\mathrm{est}}$ corresponding to prompt $y$, and $r$ represents the pretrained reward model~\cite{imagereward_xu_2023}, $\psi$ represents reward-to-loss map function implemented by ReLU. We used the DDIM~\cite{ddim_song_2021} to rapidly sample $k$ samples during the early iteration stage. This method saves 2 times the iteration step for VPSD convergence and improves the aesthetic score of the SVG by filtering out samples with low reward values in LoRA.

Our final VPSD objective is then defined by the weighted average of the three terms,
\begin{equation}
\underset{\theta}{\operatorname{min}} \; \nabla_{\theta} \mathcal{L}_{\mathrm{VPSD}} 
+ \mathcal{L}_{\mathrm{lora}} 
+ \lambda_{\mathrm{r}} \mathcal{L}_{\mathrm{reward}} 
\end{equation}
\noindent where $\lambda_{\mathrm{r}}$ indicates reward feedback strength.

\section{SVGDreamer++}
\label{sec:svgdreamer++}
\noindent In this section, we introduce the enhanced SVGDreamer++ approach. 
The original SVGDreamer exhibits two primary limitations:
(1) it may produce vector graphics with inaccurate boundaries, and its editability is limited to the object level.
(2) The number of primitives used to compose a vector graphic must be preset and remain fixed during optimization, which can lead to slow convergence or insufficient detail in the resultant vector graphics. 
To address these limitations, we introduce two improvements in SVGDreamer++. First, we propose a \textbf{H}ierarchical \textbf{I}mage \textbf{VE}ctorization (HIVE), an advanced version of SIVE, to enhance the quality of boundaries in vector graphics and extend the model's editability to both object-level and part-level (Sec.~\ref{sec:hive}). 
Second, we design an adaptive vector primitive control strategy that dynamically adjusts the number of primitives during optimization, leading to faster convergence and improved visual quality (Sec.~\ref{sec:adaptive_path_control}). 
The remaining components of SVGDreamer++ are identical to those of the original SVGDreamer.


\begin{figure}[t]
\centering
\includegraphics[width=0.6\linewidth]{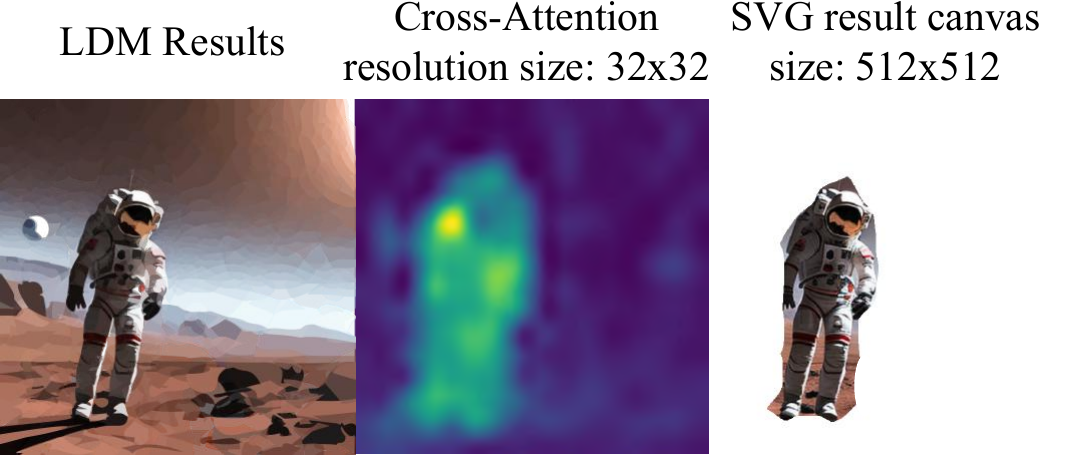}
\vspace{-1em}
\caption{
\textbf{The limitation of SIVE.} 
 When the cross attention map extracted from the LDM has a much lower resolution (\textit{e.g.}, 32x32) compared to the target vector graphic (\textit{e.g.}, 512x512), the results may have inaccurate boundaries.
} \label{fig:sive_issue}
\vspace{-1em}
\end{figure}
\subsection{HIVE: Hierarchical Image Vectorization}
\label{sec:hive}
\noindent In the SVGDreamer framework, SIVE is utilized to segregate foreground objects from the background using masks derived from the attention maps of a pre-trained diffusion model, as detailed in Sec.~\ref{sec:semantic_aware_optim}.  
However, these attention-based masks can introduce inaccuracies in boundaries during the optimization process. This issue stems from the resolution limitations of the attention features extracted from the diffusion model's cross-attention layers. As illustrated in Fig.~\ref{fig:sive_issue}, this limitation becomes evident when the resolution of the attention map is significantly lower than that of the target image. 
Furthermore, as SIVE operates at the object level, it lacks the capability to manage local or fine-grained elements, such as the helmet of a space suit.

In SVGDreamer++, we introduce a \textbf{H}ierarchical \textbf{I}mage \textbf{VE}ctorization (HIVE) approach to enhance both the quality and editability of the generated vector graphics. The core distinction between HIVE and SIVE lies in the method of generating masks, which are employed as guidance during image vectorization. HIVE utilizes segmentation priors to obtain masks, ensuring both accurate boundaries and fine-grained control.
The pipeline is shown in Fig.~\ref{fig:pipeline}. 
Specifically, HIVE adopts the primitive initialization method from SIVE, as discussed in Sec.~\ref{sec:primitive_init}. 
Subsequently, the user selects $O$ nouns from the text prompt as the trigger condition for Grounded-SAM~\cite{groundedSAM_ren_2024} to generate $O$ object-level masks.
Then, the coordinates of control points within each object are used
as conditions to drive the SAM model~\cite{SegmentAnything_kirillov_2023} to produce $F$ masks, corresponding to fine-grained details.
This results in two sets of masks: object-level masks $\{\mathcal{\hat{M}}_i\}_{i=1}^{O}$ and fine-grained masks $\{\mathcal{\tilde{M}}_j \}_{j=1}^{F}$ for individual object regions. These masks supervise the image vectorization process, as delineated in Eq.~\ref{eq:hive}.
\begin{equation}
\label{eq:hive}
\begin{split}
\mathcal{L}_{\mathrm{HIVE}} &= 
\sum_{i}^O \left( \hat{\mathcal{M}}_i \odot I - \hat{\mathcal{M}}_i \odot \bm{x} \right)^2 
\\&+ 
\sum_{i}^O \sum_{j}^F \left( \tilde{\mathcal{M}}_i^j \odot I_i - \tilde{\mathcal{M}}_i^j \odot \bm{x}_i \right)^2
\end{split}
\end{equation}
\noindent where $I$ and $I_i$ are the target image and the $i$-th object, $\{\mathcal{\hat{M}}_i\}_{i=1}^{O}$  is the set of object-level masks, with $\mathcal{\hat{M}}_i$ being the $i$-th mask predicted by Grounded-SAM, $\hat{\mathcal{M}}_i^j$ is the $j$-th fine-grained mask of the $i$-th object predicted by SAM, $\bm{x}_i=\mathcal{R}(\bm{\theta}^{\prime})$ is the $i$-th rendering.

By employing this new mask generation strategy, HIVE can effectively reduce vector path interweaving and coupling across objects or parts, significantly enhancing the visual quality and editability of the vector graphics.

\subsection{Adaptive Vector Primitives Control}
\label{sec:adaptive_path_control}
\begin{figure}[t]
\centering
\includegraphics[width=1.0\linewidth]{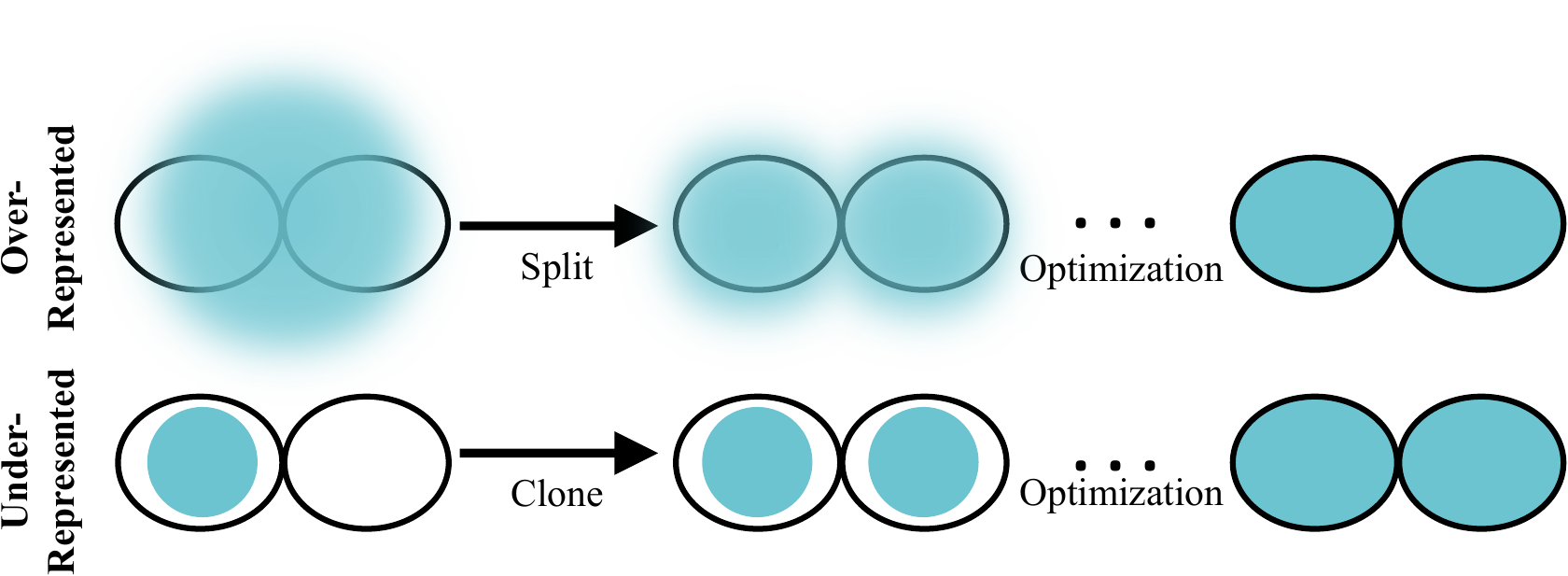}
\vspace{-2em}
\caption{
\textbf{Our Adaptive Vector Primitives Control scheme}.
Top row (Over-Represented): When a large graphic is used to represent small-scale geometry, we address this by splitting the graphic into two new graphics, each exactly half the size of the original. 
Bottom row (Under-Represented): In cases where the small-scale geometry (black outline) is not sufficiently covered, we replicate the original graphic and place the copy adjacent to the original, thus ensuring complete coverage.
} \label{fig:path_control_pipe}
\vspace{-0.5em}
\end{figure}

\noindent The number of paths significantly affects the visual quality of generated SVGs. Intuitively, complex content, such as a zebra, requires more paths than simple content, like an apple.
More paths often lead to better results by capturing more delicate details. However, an insufficient number of paths can lead to geometric feature degradation, such as missing details, while an excessive number can slow down the optimization process. Consequently, setting a \textit{proper} number of paths is a challenging task, 
and this problem remains largely unexplored. 
\begin{algorithm}[t]
\caption{Adaptive Vector Primitives Control}
\label{algo:adaptive_vec_control}
\begin{algorithmic}[1]
\Require SVG parameters $\bm{\theta}=\{ (P_i, C_i)\}_{i=1}^n$, where $C_i = \{r,g,b,\alpha\}_i$. Opacity threshold $\tau_{opacity}$, control threshold $\tau_{c}$ and area threshold $\tau_{a}$.
\While {not converged}
\State $\mathcal{L}$=ComputeLoss() \textcolor{algo_comment}{\Comment{loss computation}}
\State $\{ (P_i, C_i)\}_{i=1}^n \xleftarrow{}$ Adam($\nabla \mathcal{L}$) \textcolor{algo_comment}{\Comment{backprop \& step}}
\For { ($P_i$, $\{r,g,b,\alpha\}_i$) \text{in} $\theta$ }
    \If { $\alpha < \tau_{\text{opacity}}$} \textcolor{algo_comment}{\Comment{SVG path purning}}
    \State \text{RemovePath()}
    \EndIf
    \If { $\nabla_{\theta}\mathcal{L} > \tau_{c}$ } \textcolor{algo_comment}{\Comment{SVG path control}}
        \If { $\text{area}(P_i) > \tau_{a}$ } \textcolor{algo_comment}{\Comment{Over-Represented}}
            \State $\text{SplitPath}(P_i, \{r,g,b,\alpha\}_i)$
        \Else \textcolor{algo_comment}{\Comment{Under-Represented}}
            \State $\text{ClonePath}(P_i, \{r,g,b,\alpha\}_i)$
        \EndIf
    \EndIf
\EndFor
\EndWhile
\end{algorithmic}
\end{algorithm}
Here we introduce a novel Adaptive Vector Primitive Control strategy that can dynamically adjust the number of primitives during optimization. 
The core idea is to eliminate redundant paths and add additional paths in regions with geometric feature degradation.
As depicted in Fig.~\ref{fig:path_control_pipe}, we identify two scenarios that necessitate additional paths. In regions with complex structures, a path might cover an adequate area but be too simplistic to accurately represent the structure (termed ``Over-Represented''). In another scenario, a path might cover an insufficient area to represent the structure adequately (``Under-Represented''). Both cases can lead to geometric degradation, thus requiring more paths.
\begin{figure}[t]
\centering
\includegraphics[width=1.0\linewidth]{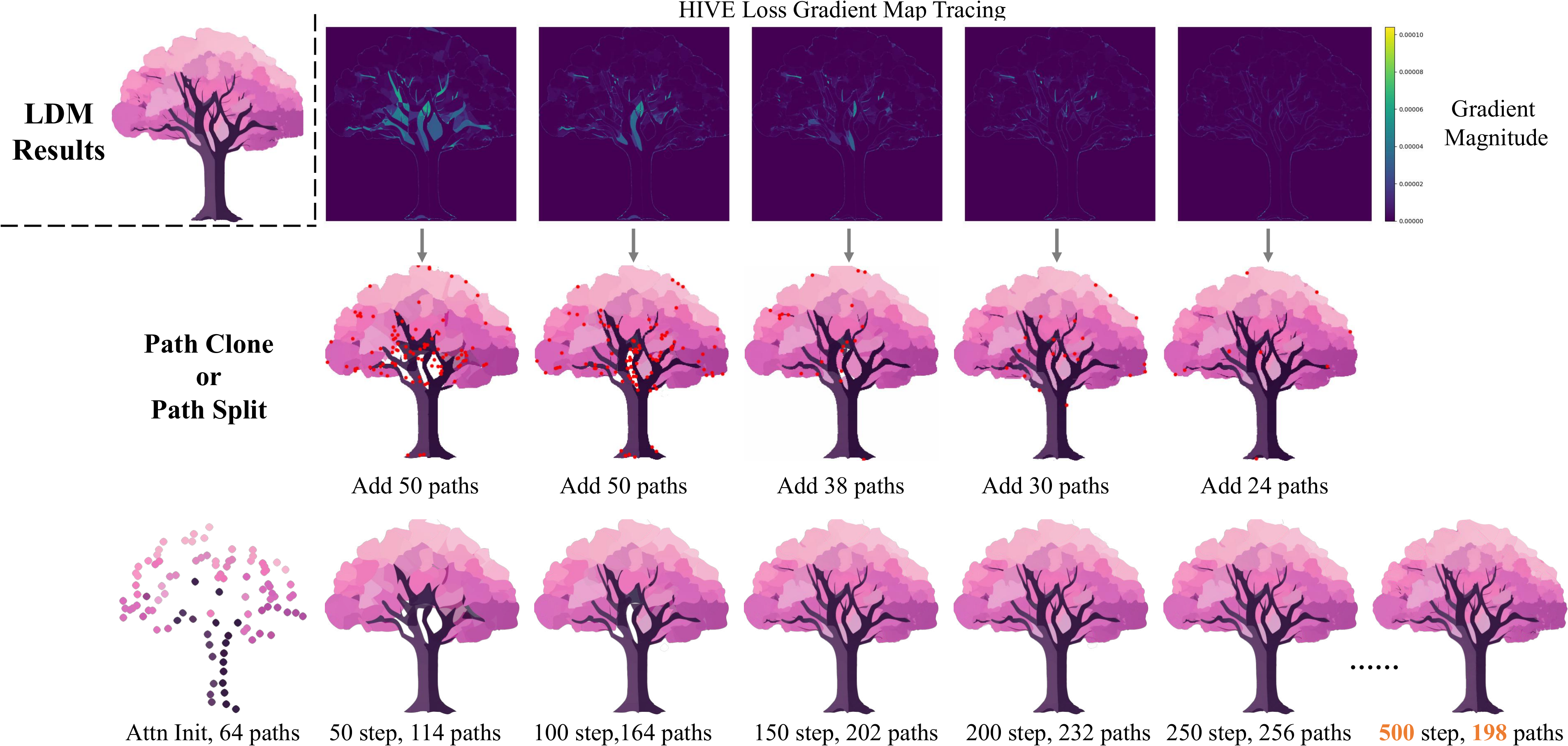}
\vspace{-2em}
\caption{
\textbf{The HIVE loss gradient map tracing.} 
The gradient map is derived by calculating the Jacobian matrix of the HIVE loss function and its gradient. Regions exhibiting higher gradient strengths suggest that the reconstruction is less effective, thereby requiring the addition of more paths. We sample the indicator points, weighted according to the gradient intensity values, to direct the vector primitive control process. All graphics containing these indicator points become the focus of our vector primitive control to enhance vectorization quality. As depicted in the 2nd row of figure, an increase in the number of strokes corresponds to the number of paths added through separation or cloning at that specific time step.
} \label{fig:hive_grad_loss_trace}
\vspace{-1em}
\end{figure}
Our \textit{Adaptive Vector Primitives Control} algorithm is detailed in Algorithm~\ref{algo:adaptive_vec_control}.
This module is designed as a plug-and-play component, capable of seamless integration into image vectorization algorithms that utilize gradient optimization. 

The algorithm includes two key components: path pruning (Lines 5 to 7) and path control (Lines 8 to 14).
Path pruning involves removing a path if its opacity falls below a certain threshold, indicating near transparency. 
Path control dynamically introduces additional paths into regions with geometric feature degradation.
For over-simple cases, a path will split into two; for over-small cases, a path will be cloned.

We identify regions requiring enhancement based on the gradient map of the loss function. 
We observe that both scenarios exhibit a large positional gradient intensity, a phenomenon likely stemming from these regions not being accurately reconstructed. 
The optimization algorithm, therefore, attempts to adjust the paths to rectify this discrepancy.
Specifically, as shown in the first row of Fig.~\ref{fig:abl_path_control}, the gradient map is derived by computing the HIVE loss and the Jacobian matrices of their gradients. We then determine the regions for improvement based on the gradient magnitude. 
Regions exhibiting higher gradient strengths suggest that the reconstruction is less effective, thereby requiring the addition of more paths. We sample the indicator points, weighted according to the gradient intensity values, to direct the vector primitive control process. All graphics containing these indicator points become the focus of our vector primitive control to enhance vectorization quality.
To enhance optimization efficiency, we vectorize the selected objects in HIVE using adaptive vector primitive control.
\section{Vector Primitives Representation}
\label{sec:various_primaries}
\noindent In addition to text prompts, we provide a variety of vector representations for style control. These vector representations are achieved by limiting primitive types and their parameters.
Users can control the art style by modifying the input text or by constraining the set of primitives and parameters.
Unlike existing text-to-image and text-to-SVG methods, we provide users a variety of flexible ways to build vector graphics, opening up potential in the field of generative vector design.
We explore six settings: 

\noindent\textbf{1) Iconography} is the most common SVG style, consisting of several paths and their fill colors. This style allows for a wide range of compositions while maintaining a minimalistic expression. We utilize closed-form Bézier curves with trainable control points and fill colors (including opacity), shown in the 1st and 2nd rows of Fig.~\ref{fig:gallery} and the top row of Fig.~\ref{fig:diverse_results}.

\noindent\textbf{2) Pixel Art} is a widely used style that draws inspiration from the low-resolution, 8-bit graphics characteristic of early video games. To emulate this style, we employ square SVG polygons with variable fill colors and opacity, enabling precise control over the pixelated aesthetic, shown in the 3rd row of Fig.~\ref{fig:gallery} and 2nd row of Fig.~\ref{fig:diverse_results}.

\noindent\textbf{3) Low-Poly Art} involves the deliberate cutting and arrangement of simple geometric shapes according to the modeling principles of objects. To achieve this style, we utilize square SVG polygons with trainable control points and variable fill colors (including opacity), which enables precise control over the composition and aesthetic of the low-poly representation, shown in the 4th row of Fig.~\ref{fig:gallery} and 3rd row of Fig.~\ref{fig:diverse_results}.

\noindent\textbf{4) Painting Style} vector art seeks to replicate a painter's brush strokes within the vector domain. This is achieved through the use of open-form Bézier curves with trainable control points, variable stroke color (including opacity), and adjustable stroke width, allowing for precise emulation of traditional painting techniques, shown in the 5th row of Fig.~\ref{fig:gallery} and 4th row of Fig.~\ref{fig:diverse_results}.

\noindent\textbf{5) Sketching} employs black strokes to delineate objects, serving as a method to convey information with minimalistic expression. To replicate this style, we utilize open-form Bézier curves with trainable control points and adjustable opacity, allowing for precise control over the sketch-like appearance, shown in the 5th row of Fig.~\ref{fig:gallery} and 6th row of Fig.~\ref{fig:diverse_results}.

\noindent\textbf{6) Ink and Wash Painting} is a traditional Chinese art form characterized by the use of varying concentrations of black ink to create nuanced and expressive imagery. To emulate this style in our work, we employ open-form Bézier curves with trainable control points, adjustable opacity, and variable stroke widths, enabling precise control over the rendering of ink-like effects, shown in the 5th row of Fig.~\ref{fig:gallery} and 5th row of Fig.~\ref{fig:diverse_results}.

%% file: sec/4_experiment.tex
\section{Experiments}
\label{sec:experiments}
\begin{table*}[t]
\caption{
\textbf{Quantitative Comparison of SVGDreamer++ v.s. state-of-the-art Text-to-SVG Methods.}
${\dag}$: our reproduced results.
}
\vspace{-1em}
\resizebox{1.0\linewidth}{!}{
\begin{tabular}{l|cccccc}
\toprule
Method / Metric &FID~\cite{FID_Heusel_2017}$\downarrow$&PSNR~\cite{PSNR_Hore_2010}$\uparrow$&CLIP Score~\cite{CLIP_radford_2021}$\uparrow$&BLIP Score~\cite{blip_li_2022}$\uparrow$ &Aesthetic~\cite{aesthetic_christoph_2022}$\uparrow$&HPS~\cite{HPS_Wu_2023}$\uparrow$\\
\midrule
CLIPDraw~\cite{clipdraw_frans_2022} &131.65&8.35&0.2486&0.3933&3.9803&0.2347\\
Evolution~\cite{evolution_tian_2022} &161.43&3.70&0.1932&0.3450&4.0845&0.1955\\
$\text{DiffSketcher}$\cite{diffsketcher_xing_2023} &77.30&6.75&0.2402&0.4185&4.1562&0.2423\\
$\text{VectorFusion(scratch)}^{\dag}$~\cite{vectorfusion_jain_2023} &80.76&6.33&0.2298&0.3803&4.5165&0.2334\\
$\text{VectorFusion}^{\dag}$~\cite{vectorfusion_jain_2023} &69.22&8.01&0.2720&0.4291&4.9845&0.2450 \\
\midrule
\textbf{SVGDreamer}(scratch) &39.30&10.51&0.2988&0.4335&5.2825&0.2559\\
\rowcolor{maroon!10}
\textbf{SVGDreamer}&30.10&14.54&0.3001&0.4623&5.5432&0.2685\\
\rowcolor{maroon!40}
\textbf{SVGDreamer++}&\textbf{22.13}&\textbf{15.80}&\textbf{0.3093}&\textbf{0.4701}&\textbf{5.6124}&\textbf{0.2760}\\
\bottomrule
\end{tabular}
} \label{tab:quantitative}
\end{table*}
\begin{figure*}[ht]
\centering
\includegraphics[width=1\linewidth]{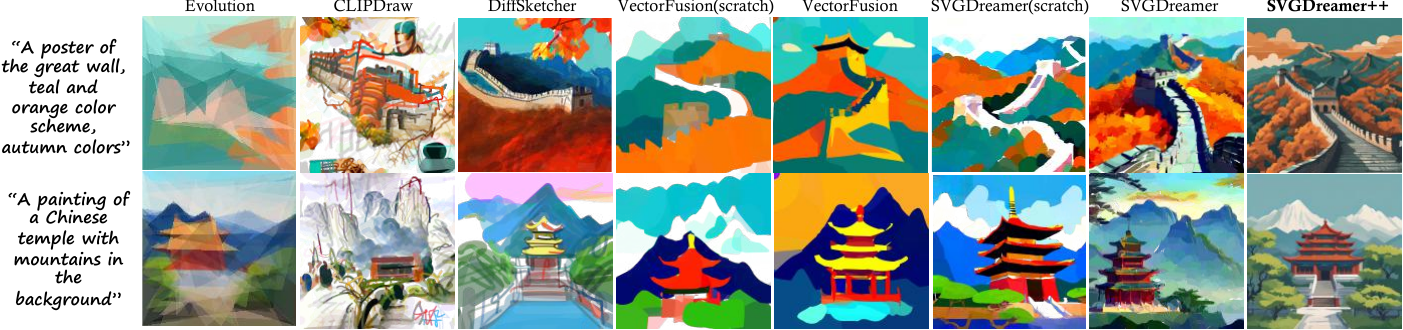}
\vspace{-2em}
\caption{
\textbf{Qualitative comparison of SVGDreamer++ vs. the state-of-the-art Text-to-SVG methods.}
Note that DiffSketcher was originally designed for vector sketch generation; therefore, we re-implemented it to generate RGB vector images. SVGDreamer++ is capable of composing complex and highly detailed vector images, particularly in representing tree elements and architectural details.
} \label{fig:compare_methods}
\vspace{-1em}
\end{figure*}
\noindent\textbf{Overview.} 
In this section, we first explain the dataset and evaluation metrics we used, as well as the implementation details of our experiments. We then provide experimental results to demonstrate the effectiveness of our proposed method. 
Specifically, Section~\ref{sec:qualitative_and_quantitative} offers a qualitative (Sec.~\ref{sec:qualitative}) and quantitative (Sec.~\ref{sec:quantitative}) comparison with state-of-the-art methods, accompanied by a flowchart (Sec.~\ref{sec:editability}) illustrating the SVG editing process. 
Section~\ref{sec:abl_study} presents ablation studies and analytical results for deeper insights.
Section~\ref{sec:application} demonstrates the practical applications of the proposed SVGDreamer++ in vector design, particularly in designing posters (Sec.~\ref{sec:poster_desgin}) and generating vector assets (Sec.~\ref{sec:vector_assets}).

\noindent\textbf{Dataset.}
Current text-to-SVG approaches perform well on prompts with a single simple portrait object but struggle with prompts that include environmental surroundings or multiple objects due to inaccurate 2D supervision. To evaluate these methods, we design three prompt sets: Single object, Single object with surroundings, and Multiple objects. The Single object set establishes a baseline, while the other two sets increase complexity. We then use these three prompt sets to conduct a thorough evaluation of text-to-SVG methods. 

\noindent\textbf{Evaluation Metrics.}
To evaluate our proposed method and baseline methods, we employed six quantitative indicators across four dimensions: 
(1) Visual quality of the generated SVGs, assessed by \textbf{FID} (Fréchet Inception Distance)~\cite{FID_Heusel_2017}; (2) Fidelity of color representation, evaluated by \textbf{PSNR} (Peak Signal-to-Noise Ratio)~\cite{PSNR_Hore_2010}; (3) Alignment with the input text prompt, assessed by \textbf{CLIP score}~\cite{CLIP_radford_2021} and \textbf{BLIP score}~\cite{blip_li_2022}, and (4) Aesthetic appeal of the generated SVGs, measured by \textbf{Aesthetic score}~\cite{aesthetic_christoph_2022} and \textbf{HPS} (Human Preference Score)~\cite{HPS_Wu_2023}. 

\label{sec:implement}
\noindent \textbf{Implementation Details.} 
In our implementation, we leverage the pre-trained Stable Diffusion~\cite{ldm_Rombach_2022}. For SVG parameter optimization $\theta=\{P_i, C_i\}_{i=1}^n$, we use the Adam optimizer with settings $\beta_1=0.9$, $\beta_2=0.9$, $\epsilon=1e-6$. 
We use a learning rate warm-up strategy where the control point learning rate starts at 0.01 and increases to 0.9 over the first 50 iterations, followed by an exponential decay from 0.8 to 0.4 over the subsequent 650 iterations, totaling 700 iterations.
The color learning rate is set to 0.1 and the stroke width learning rate to 0.01.
For the training of LoRA~\cite{lora_hu_2022} parameters, We adopt the AdamW optimizer with parameters $\beta_1=0.9$, $\beta_2=0.999$, $\epsilon=1e-10$, and $lr=1e-5$. 
In the HIVE experiment, to counteract the vacant background regions caused by segmentation, we integrate the LaMa model~\cite{lama_suvorov_2021} to fill these areas before processing with HIVE for vectorization.
In most experiments, we set the particle number $k$ to 6, which means that six particles simultaneously participate in the VPSD update.
To ensure diversity and fidelity in the synthesized SVGs while preserving rich details, we set the guidance scale of the Classifier-free Guidance~\cite{classifierfree_2022_ho} (CFG) to 7.5.
During the optimization process of SVGDreamer++, we introduce the adaptive path control algorithm at the 200th iteration, and subsequently 
every 25 iterations. The opacity threshold $\tau_{\text{opacity}}$ is set to 0.05, and the control threshold $\tau_{\text{c}}$ is adjusted to 1e-5 to ensure precision in vectorization. For a canvas size of 1024, the area threshold $\tau_{\text{a}}$ is defined as 20,000, and for a canvas size of 768, this threshold is set to 10,000. These settings are optimized to balance computational efficiency and detail accuracy in the resulting vector graphics.
\begin{figure*}[t]
\centering
\includegraphics[width=1.0\linewidth]{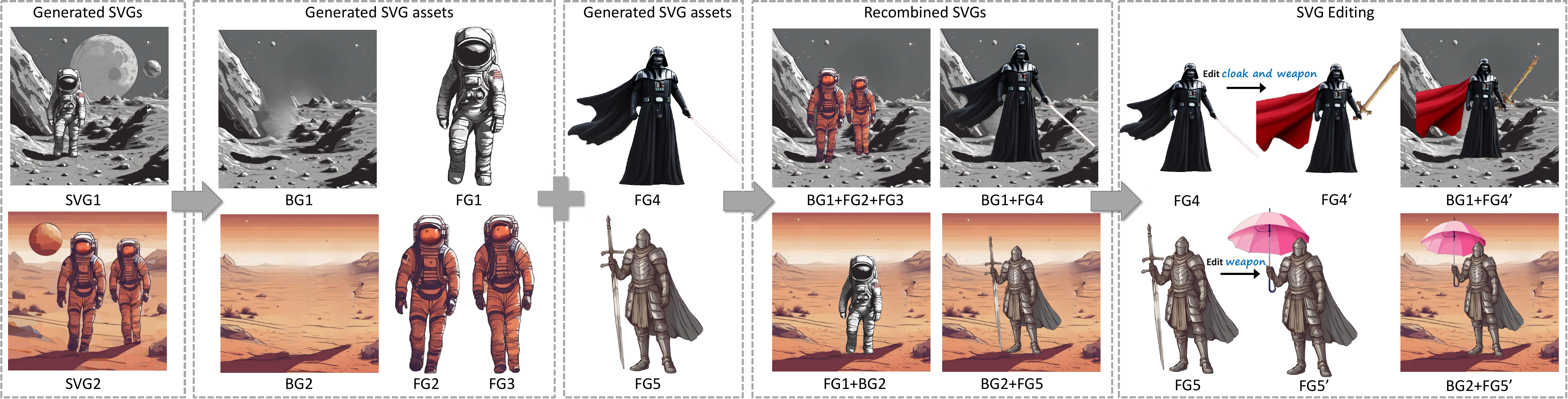}
\vspace{-2em}
\caption{
\textbf{The editability of SVGDreamer++ results.} 
Our process initiates with the examination of two SVGs generated by SVGDreamer++ ($\text{SVG1}$ and $\text{SVG2}$), where we first illustrate the decoupled vector elements at the object level ($\text{BG1}, \text{FG1}$, \text{BG2}, \text{FG2} and \text{FG3}).  
Subsequently, we generate two new vector objects using the SVGDreamer++ framework ($\text{FG4}$ and $\text{FG5}$).
As depicted in the fourth dotted box, our methodology facilitates object-level editing ($\text{BG1}$+$\text{FG2}$+$\text{FG3}$, $\text{FG1}$+$\text{BG2}$). 
Moreover, the fifth dotted box exemplifies the capability to edit individual objects, including local elements of vector objects.
For example, Darth Vader's lightsaber is altered to a gold longsword, and his black cloak is modified to a red cloak ($\text{FG4} \rightarrow \text{FG4}^{\prime}$). Another example is the transformation of a large sword into a pink umbrella ($\text{FG5} \rightarrow \text{FG5}^{\prime}$). Finally, we put the edited vector object back into the background (
$\text{BG1}$+$\text{FG4}^{\prime}$, $\text{BG2}$+$\text{FG5}^{\prime}$).
} \label{fig:editable}
\vspace{-1em}
\end{figure*}
\subsection{Qualitative and Quantitative Evaluation}
\label{sec:qualitative_and_quantitative}
\subsubsection{Qualitative Results}
\label{sec:qualitative}
Figure~\ref{fig:compare_methods} presents a qualitative comparison between SVGDreamer++ and existing text-to-SVG methods~\cite{evolution_tian_2022,clipdraw_frans_2022,diffsketcher_xing_2023,vectorfusion_jain_2023,svgdreamer_xing_2023}. 
Notably, VectorFusion(scratch)~\cite{vectorfusion_jain_2023} and SVGDreamer(scratch)~\cite{svgdreamer_xing_2023} represent variants of each method that omit the image vectorization step, focusing solely on optimization with SDS or VPSD, respectively.

Our observations are as follows:
(1) Results from CLIP-based methods, including CLIPDraw~\cite{clipdraw_frans_2022} and Evolution~\cite{evolution_tian_2022}, fail to effectively match the input text prompts. This can be explained by that CLIP-based methods lack the generative capacity to accurately reproduce text descriptions.
(2) Diffusion model-based methods such as DiffSketcher~\cite{diffsketcher_xing_2023} and VectorFusion~\cite{vectorfusion_jain_2023} demonstrate the ability to generate SVGs that are faithful to text prompts.  However, the performance of DiffSketcher is less satisfactory as it is primarily designed for sketch generation.
Furthermore, the use of SDS in VectorFusion results in vector shapes that appear overly smooth and colors that are overly saturated. This issue becomes more evident when comparing the results of VectorFusion(scratch) with SVGDreamer(scratch), essentially highlighting the differences between SDS and VPSD.
We hypothesize that the random timestep sampling technique used in SDS introduces lower-quality, distorted shapes during optimization, which degrades the overall quality of the samples.
(3) Compared to SDS-based methods~\cite{vectorfusion_jain_2023,diffsketcher_xing_2023}, both SVGDreamer and SVGDreamer++ which utilize VPSD, effectively address issues such as shape over-smoothing and color over-saturation.
This improvement is due to VPSD's ability to promote sample diversity by separately learning the distributions of control points and colors in vector graphics.
Additionally, ReFL is introduced in each iteration to assess the quality of sample reconstruction, aligning the results more closely with human aesthetics.
(4) With the newly proposed HIVE module and adaptive vector primitive control strategy, SVGDreamer++ achieves SVGs with enhanced visual quality compared to SVGDreamer.

\subsubsection{Quantitative Results}
\label{sec:quantitative}
\noindent 
Table~\ref{tab:quantitative} compares our proposed method with baseline methods using six quantitative indicators across four dimensions. The results are aligned with the qualitative results discussed in the previous section. 
Specifically, 
CLIPDraw~\cite{clipdraw_frans_2022} and Evolution~\cite{evolution_tian_2022} achieve an FID of 131.65 and 161.43, respectively, which are significantly higher than those of other methods. This indicates that these two methods struggle to produce high-quality SVGs. 
In contrast, VectorFusion~\cite{vectorfusion_jain_2023} and DiffSketcher~\cite{diffsketcher_xing_2023}, both are diffusion model-based methods, show improved results.
VectorFusion achieves an FID of 69.22 and a PSNR of 8.01, while DiffSketcher achieves an FID of 77.30 and a PSNR of 6.75. 
Although these values represent an improvement over CLIPDraw and Evolution, their FIDs (and PSNRs) are still relatively high (and low), suggesting that the visual quality of their output SVGs is not entirely satisfactory.
Furthermore, the comparison of SVGDreamer(scratch)~\cite{svgdreamer_xing_2023} and VectorFusion(scratch)~\cite{vectorfusion_jain_2023} in terms of PSNR highlights the effectiveness of VPSD in relieving the issue of color over-saturation.
Both SVGDreamer~\cite{svgdreamer_xing_2023} and SVGDreamer++ achieve lower FIDs, indicating that their output SVGs possess substantially higher visual quality. 
With the incorporation of the ReFL module, these methods also achieve high scores in aesthetic score and HPS.
Finally, SVGDreamer++, the enhanced version of SVGDreamer, achieves the best performance across all evaluated metrics, with a remarkable FID of 22.13 and a PSNR of 15.80. The improvements in the CLIP Score, BLIP Score, Aesthetic Score, and HPS further underscore the superiority of SVGDreamer++ in generating SVGs that are more aligned with text prompts and human preferences.
\subsubsection{Editability}
\label{sec:editability}
With our newly proposed HIVE module (Sec.~\ref{sec:hive}), SVGDreamer++ is capable of generating high-quality vector graphics that are editable at both the \textit{object-level} and \textit{part-level}. 
As shown in Fig.~\ref{fig:editable}, this capability empowers users to efficiently reuse synthesized vector elements and create new vector compositions. 
Two SVGs generated by SVGDreamer++ ($\text{SVG1}$ and $\text{SVG2}$), can be decoupled at the object level into components including $\text{BG1}, \text{FG11}$, \text{BG2}, \text{FG2} and \text{FG3}.  
These foreground objects and background elements can be recombined to form new SVGs, as demonstrated in the fourth box of the Fig.~\ref{fig:editable} ($\text{BG1}$+$\text{FG2}$+$\text{FG3}$, $\text{FG1}$+$\text{BG2}$, $\text{BG1}$+$\text{FG4}$ and $\text{BG2}$+$\text{FG5}$).
Furthermore, as shown in the fifth box, local elements can also be editied. For example, the cloak of the character can be changed from a black one to a red one, while his weapon is changed from a lightsaber to a golden longsword. 
Finally, after editing, we put the character back into the background.
In summary, this example demonstrates that the results generated by SVGDreamer++ are editable at both the object level and the local level.

\begin{figure}[t]
\centering
\includegraphics[width=1\linewidth]{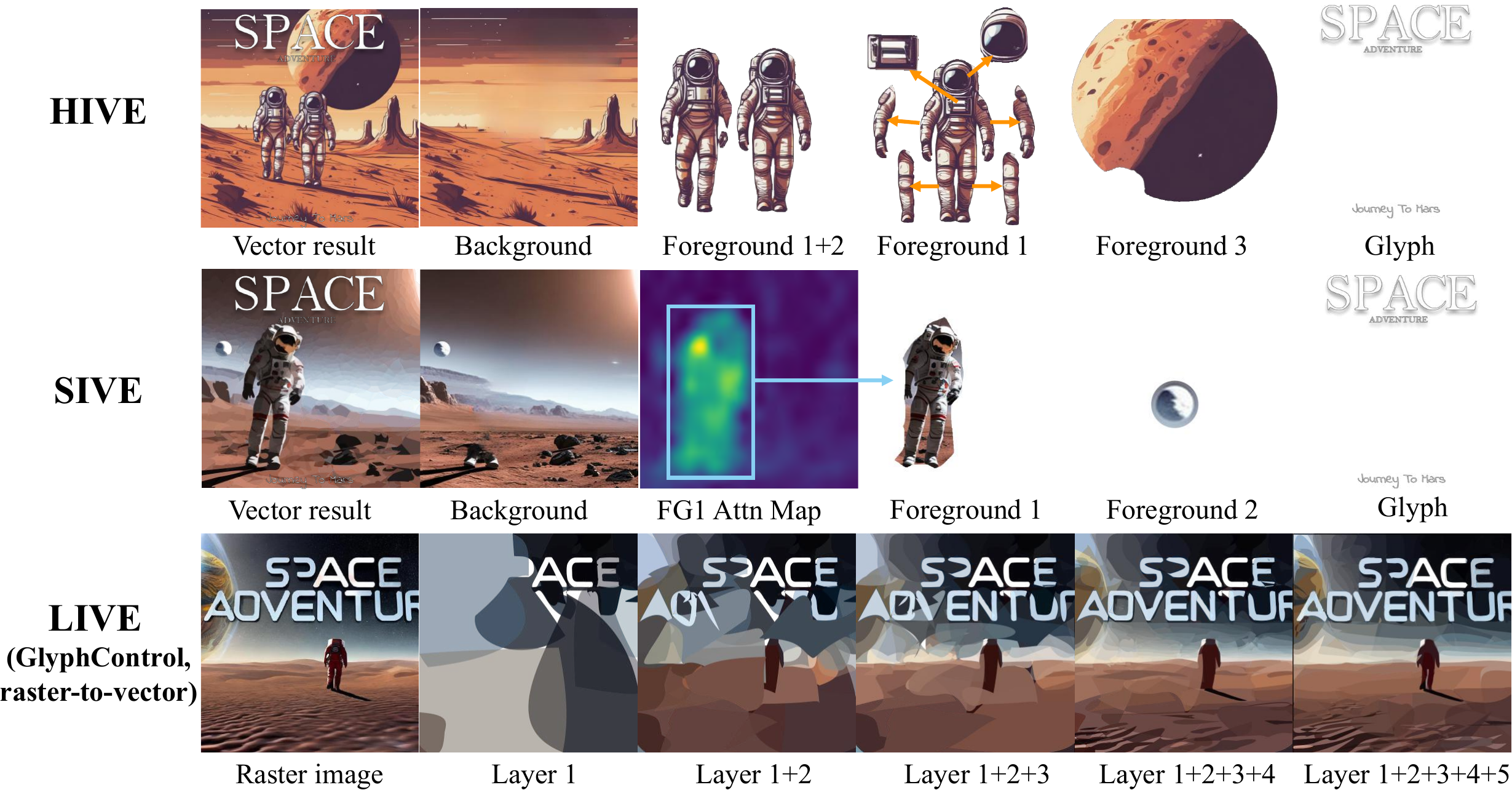}
\vspace{-2em}
\caption{
\textbf{Comparison of HIVE vectorization results with LIVE and SIVE.}
The 1st row represents HIVE, which not only decouples vector elements between objects but also separates object composition into distinct components.
The 2nd row represents SIVE, which manages the vectorization of objects from the attention map and decouples vector elements solely between objects. Nonetheless, the inherent resolution limitations of attention diagrams lead to boundary errors in vector elements.
The 3rd row represents LIVE~\cite{LIVE_Ma_2022}, we follow the protocol outlined in VF~\cite{vectorfusion_jain_2023}, which represents a vector image with 160 paths distributed across five layers, with 32 paths in each layer. The results generated by LIVE obscure the distinctions between fonts and content, as well as between individual objects, thereby diminishing its capacity for precise editing.
} \label{fig:abl_live_vs_sive_vs_five}
\vspace{-1em}
\end{figure}
\subsection{Ablation Study}
\label{sec:abl_study}
\subsubsection{HIVE v.s. SIVE v.s. LIVE}
\label{sec:abl_five_sive_live}
\begin{figure}[t]
\centering
\includegraphics[width=1.0\linewidth]{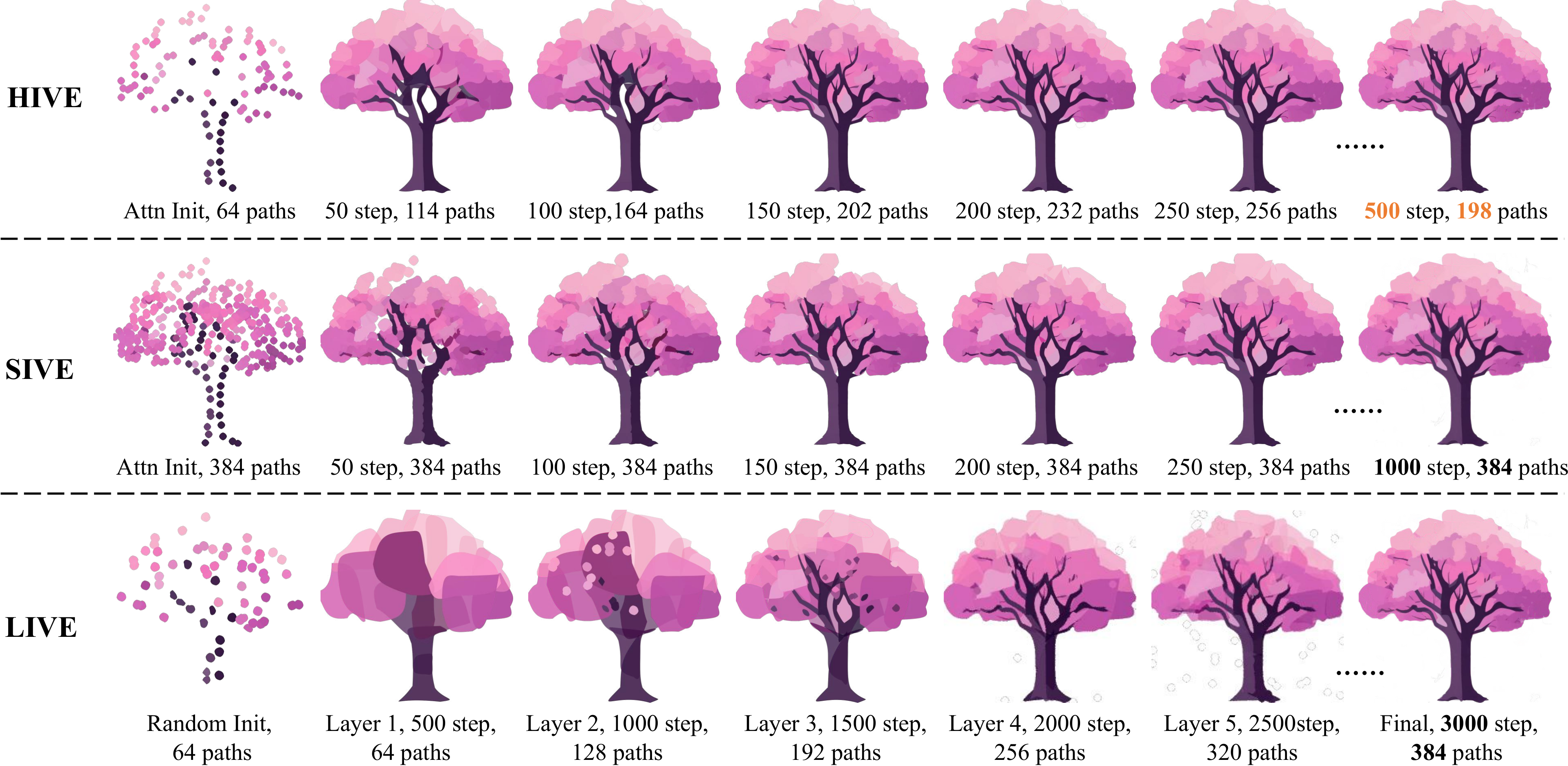}
\vspace{-2em}
\caption{
\textbf{Comparison of HIVE vectorization process with LIVE and SIVE.}
HIVE outperforms LIVE in terms of accuracy for vector path control by strategically utilizing gradient strength to guide path adjustments and accurately determine control points for operations such as cloning and splitting.
Moreover, unlike SIVE, which relies on a predetermined number of strokes, HIVE dynamically optimizes the number of paths during the process, as illustrated in the figure above, where SIVE's initial path count is identical to the final path count achieved by LIVE.
} \label{fig:abl_path_control}
\end{figure}
Figure~\ref{fig:abl_live_vs_sive_vs_five} presents a comparative analysis of the three vectorization techniques.
As illustrated in the 3rd row of Fig.~\ref{fig:abl_live_vs_sive_vs_five}, LIVE~\cite{LIVE_Ma_2022} encounters considerable challenges in accurately capturing and differentiating discrete subject elements within images. This frequently results in the overlay of identical paths across varying visual subjects, such as fonts, astronauts, and backgrounds, leading to significant path confusion.
When addressing complex vector graphic tasks that involve multiple paths, LIVE often produces hierarchical path overlays across different objects. This introduces additional complexity into SVG representations, thereby complicating subsequent editing processes.
The 2nd row of Fig.~\ref{fig:abl_live_vs_sive_vs_five} demonstrates that SIVE (Sec.~\ref{sec:SIVE}) assigns paths to vector objects, facilitating object-level vectorization. However, the limitations in resolution of cross-attention maps contribute to inaccuracies in boundary delineation, with vector boundaries for elements such as astronauts and planets occasionally blending into the background.
The HIVE (Sec.~\ref{sec:hive}) methodology proposed in this paper effectively mitigates these limitations by offering precise supervisory signals for vector objects throughout the optimization process.
Moreover, HIVE provides advanced support for both object-level and part-level vectorization, thereby enabling detailed local editing of vector objects and extending the capabilities beyond those of previous approaches.

\begin{figure}[t]
\centering
\includegraphics[width=1.0\linewidth]{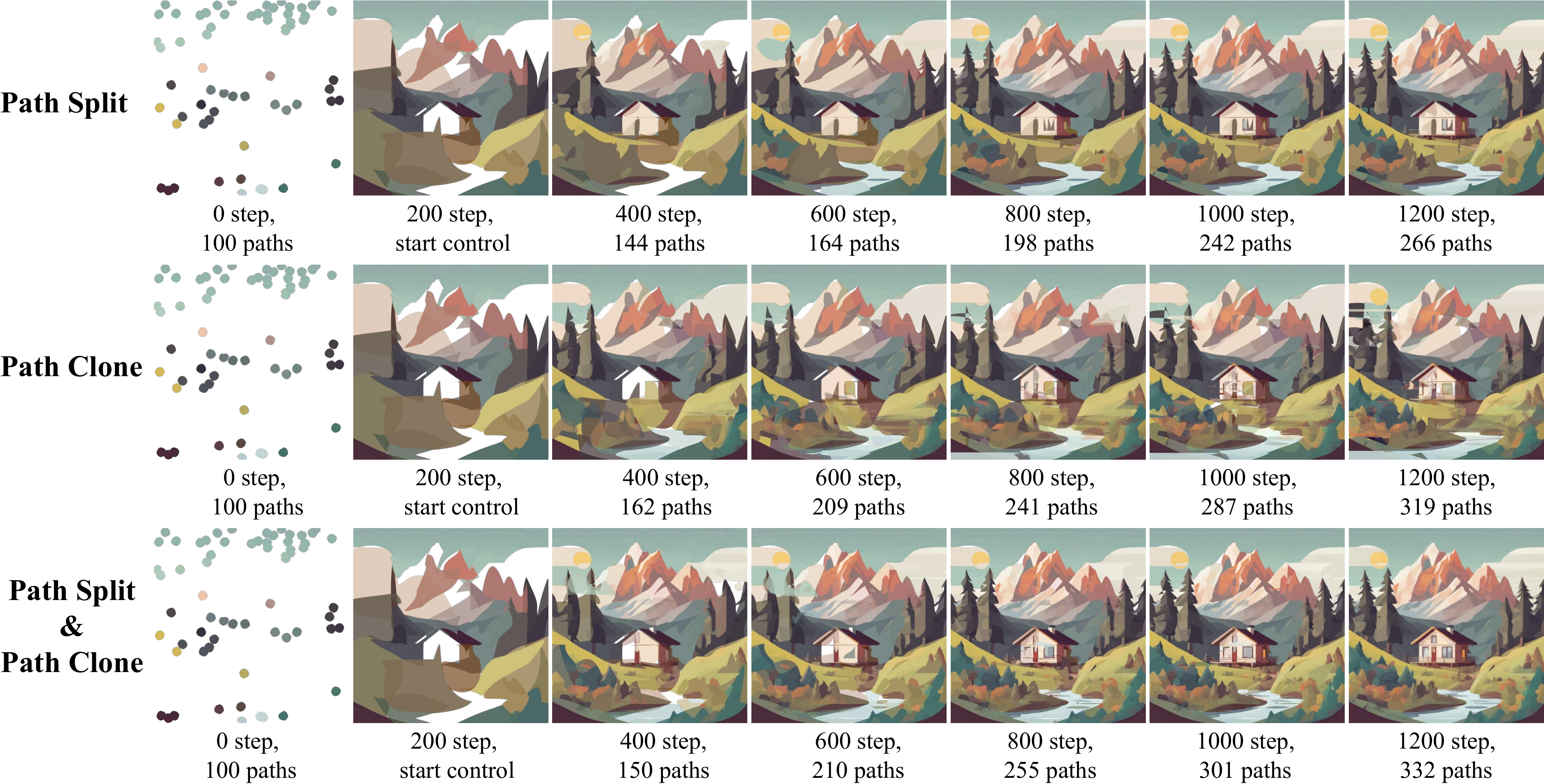}
\vspace{-2em}
\caption{
\textbf{Our Adaptive Vector Primitives Control behavior}.
To illustrate the behavior of Adaptive Vector Primitives Control, we visualize its vectorization process.
Employing the same random initialization to generate 100 paths, we implement the Adaptive Vector Primitives Control algorithm beginning at 200 steps. 
The first line illustrates the application of Path Split, which significantly enhances reconstruction details, particularly when the path covers extensive areas. 
The second line exemplifies Path Cloning, which further refines the details. 
The third line integrates both techniques within our comprehensive Adaptive Vector Primitives Control. 
Notably, all three methods incorporate path pruning to optimize the vectorization process.
} \label{fig:path_control_process}
\vspace{-1em}
\end{figure}

\subsubsection{The Impact of Adaptive Vector Primitives Control}
\label{sec:impact_of_adaptive_path_control}
As shown in Fig.~\ref{fig:abl_path_control}. We input text prompt ``A tree, color palette: light pink and purple. minimalism. flat 2d" into the Latent Diffusion Model (LDM)~\cite{ldm_Rombach_2022} to generate raster images (the first one in the first row) and visualize the processes involved in three distinct image vectorization methods.
The first line illustrates the gradient visualization of the HIVE loss functions.
By leveraging the gradient intensity, we guide path control and identify control points for cloning or splitting (The second row in the figure).   
In comparison to LIVE, HIVE offers enhanced precision in controlling vector paths.
Furthermore, compared to SIVE, HIVE dynamically adjusts the number of paths without the need for pre-specification.
VectorFusion adapts LIVE, which employs a coarse-to-fine strategy. This approach first uses large paths to delineate rough shapes and then adds smaller paths to express details. However, when the initial large paths are placed incorrectly, more smaller paths are required to compensate in later layers, potentially slowing down the optimization process.

As shown in Fig.~\ref{fig:path_control_process}, we illustrate the behavior of \textit{Adaptive Vector Primitives Control} by visualizing the vectorized representation of its components. 
Path split is employed to divide a graph that covers a large area into two parts, thus enabling a more detailed representation of the target graph. Path clone, on the other hand, creates a duplicate of the path to further refine its details. When applied to graphics of varying areas, the combination of these two methods proves to be more efficient.

\noindent For the ablation experiments of VPSD (\cref{sec:sds_compare},~\cref{sec:vector_particle_impact}, and~\cref{sec:ReFL_impact}), we utilize the results from the SVGDreamer version to assess its impact and performance.
\begin{figure}[t]
\centering
\includegraphics[width=1.0\linewidth]{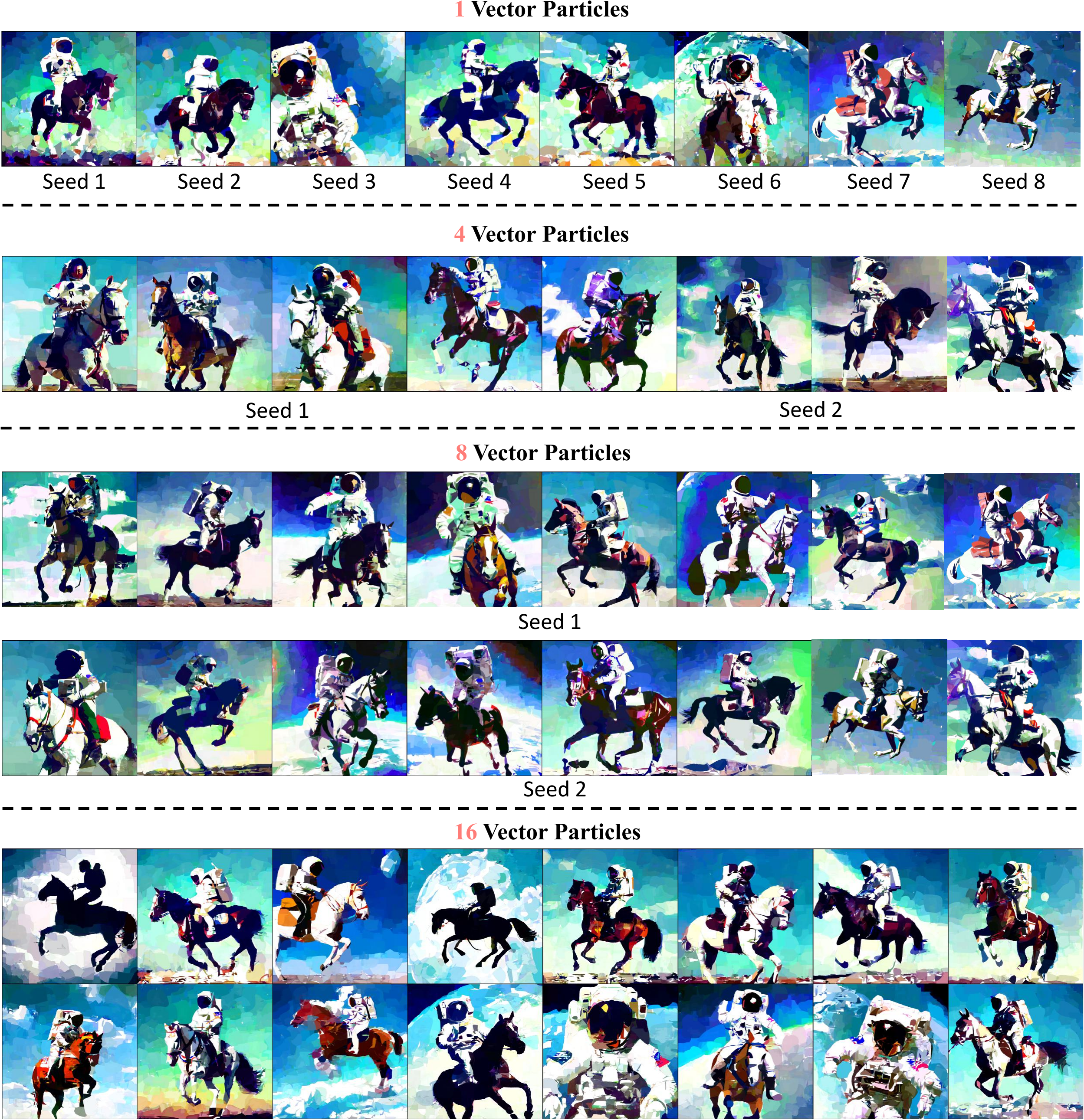}
\vspace{-2em}
\caption{
\textbf{Effects of the number of vector particles in VPSD}.
The diversity of the generated results is slightly larger as the number of particles increases.  The quality of generated results is not significantly affected by the number of particles. The prompt is ``A photograph of an astronaut riding a horse''.
} \label{fig:supp_particles}
\vspace{-1em}
\end{figure}
\subsubsection{VPSD v.s. LSDS v.s. ASDS}
\label{sec:sds_compare}
The development of text-to-SVG~\cite{vectorfusion_jain_2023,diffsketcher_xing_2023} was inspired by DreamFusion~\cite{dreamfusion_poole_2023}, but the resulting vector graphics have limited quality and exhibit a similar over-smoothness as the DreamFusion reconstructed 3D models. 
The main distinction between ASDS~\cite{diffsketcher_xing_2023} and LSDS~\cite{vectorfusion_jain_2023,wordasimg_Iluz_2023} lies in the augmentation of the input data.
As demonstrated in Tab~.\ref{tab:quantitative} and Fig.~\ref{fig:compare_methods}, our approach demonstrates superior performance compared to the SDS-based approach in terms of FID. This indicates that our method is able to maintain a higher level of diversity without being affected by mode-seeking disruptions. Additionally, our approach achieves a higher PSNR compared to the SDS-based approach, suggesting that our method avoids the issue of supersaturation caused by averaging colors.
\subsubsection{The Impact of the Number of Vector Particles}
\label{sec:vector_particle_impact}
\noindent We investigate the impact of the number of particles on the generated results. We vary the number of particles in 1, 4, 8, 16 and analyze how this variation affects the outcomes. 
As shown in Fig.~\ref{fig:supp_particles}, the diversity of the generated results is slightly larger as the number of particles increases. Meanwhile, the quality of generated results is not significantly affected by the number of particles. 
Considering the high computation overhead associated with optimizing vector primitive representations and the limitations imposed by available computation resources, we limit our testing to a maximum of 6 particles.

\subsubsection{The Impact of Reward Feedback Learning (ReFL)}
\label{sec:ReFL_impact}
\begin{table}[t]
\caption{
\textbf{Effects of introducing the Reward Learning (Sec.~\ref{sec:vpsd}) in VPSD.}
We set the number of vector particles to 1. The experiment was conducted on a single NVIDIA A800 GPU.
}
\centering
\vspace{-1em}
\resizebox{1.0\linewidth}{!}{
\begin{tabular}{c|c|c|c|c}
\toprule
Method&Canvas Size&Path Number&Iteration Steps&Time(min:sec)\\
\midrule
W/O ReFL &224*224&128&500&13m15s\\
\midrule
W ReFL &224*224&128&300&6m45s \\
\midrule
W/O ReFL &600*600&256&500&14m21s \\
\midrule
W ReFL &600*600&256&300&7m21s \\
\bottomrule
\end{tabular}
} \label{tab:refl}
\end{table}
\begin{figure}[t]
\centering
\includegraphics[width=1\linewidth]{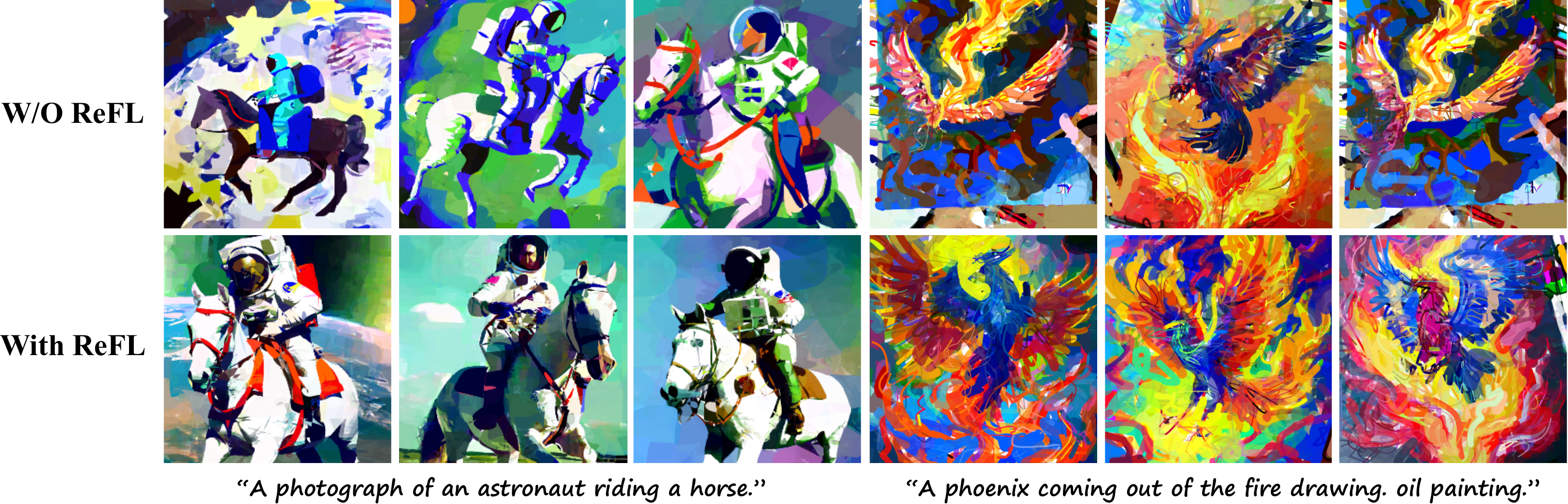}
\vspace{-2em}
\caption{
\textbf{Effects of the Reward Learning in VPSD}. When employing Reward Learning, the visual quality of the generated results is significantly enhanced.
} \label{fig:supp_refl}
\vspace{-1em}
\end{figure}
\begin{figure*}[t]
\centering
\includegraphics[width=0.98\linewidth]{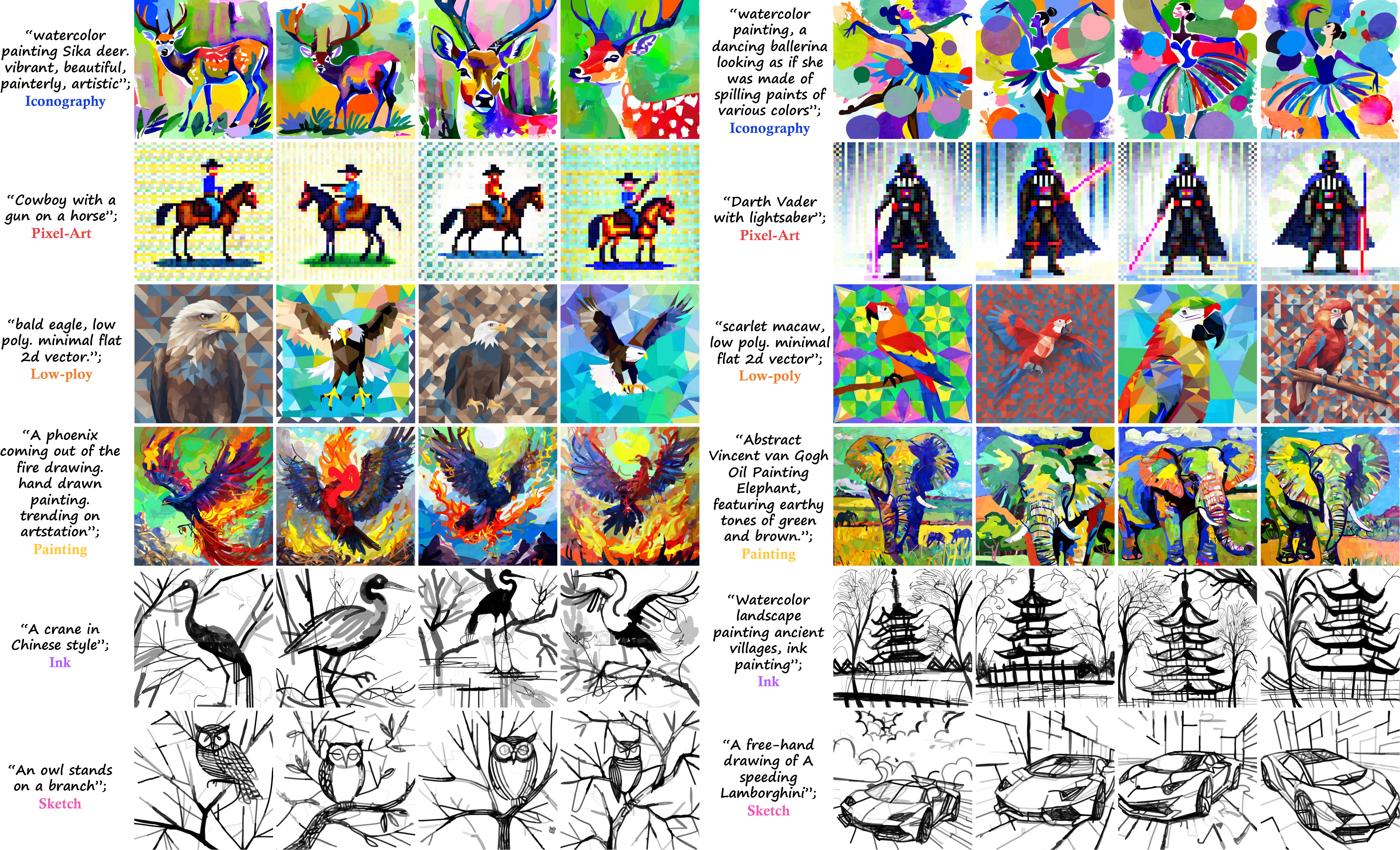}
\vspace{-1em}
\caption{
\textbf{SVG diversity generated by SVGDreamer++}. We set the number of vector particles in SVGDreamer++ to 4 to synthesize diverse results. The results show that our method can maintain SVG quality and has variety.
} \label{fig:diverse_results}
\vspace{-1em}
\end{figure*}
\noindent In~\cite{prolificdreamer_wang_2023}, only selected particles update the LoRA network in each iteration. However, this approach neglects the learning progression of LoRA networks, which are used to represent variational distributions. These networks typically require numerous iterations to approximate the optimal distribution, resulting in slow convergence. Unfortunately, the randomness introduced by particle initialization can lead to early learning of sub-optimal particles, which adversely affects the final convergence result.
In VPSD, we introduce a Reward Feedback Learning (ReFL) method. This method leverages a pre-trained reward model~\cite{imagereward_xu_2023} to assign reward scores to samples collected from LoRA model. Then LoRA model subsequently updates from these reweighted samples.
As indicated in Tab.~\ref{tab:refl}, this led to a significant reduction in the number of iterations by almost 50\%, resulting in a 50\% decrease in optimization time.
And improves the aesthetic score of the SVG by filtering out samples with low reward values in LoRA. 
Filtering out samples with low reward values, as demonstrated in Tab.~\ref{tab:quantitative}, enhances the aesthetic score of the SVG.
The visual improvements brought by ReFL are illustrated in Fig.~\ref{fig:supp_refl}.

\subsubsection{SVG Diversity Generation}
\label{sec:diverse_style_svg_gen}
\noindent As depicted in Fig.~\ref{fig:diverse_results}, we offer a diverse array of vector representations to facilitate style control, extending beyond mere text prompts to include constraints on primitive types and their parameters.
In Sec.~\ref{sec:various_primaries}, we delineate six vector styles, each characterized by unique combinations of vector primitives. This diverse definition enables a more flexible and precise representation of styles in the domain of vector graphics. Users can manipulate the artistic style by adjusting the input text or restricting the set of primitives and their associated parameters. Distinct from existing text-to-image and text-to-SVG methods~\cite{clipdraw_frans_2022,Clipasso_vinker_2022,diffsketcher_xing_2023,T2VecNeualPath_zhang_2024,NIVeL_thamizharasan_2024}, our approach affords users flexible and varied means to generate vector graphics, thereby broadening the scope of generative vector design. Notably, VF~\cite{vectorfusion_jain_2023} initially offers three vector styles—iconography, sketch, and pixel-art. We have expanded this repertoire to six by incorporating ink-and-wash, low-polygon, and painting styles.
\begin{figure}[h]
\centering
\includegraphics[width=1.0\linewidth]{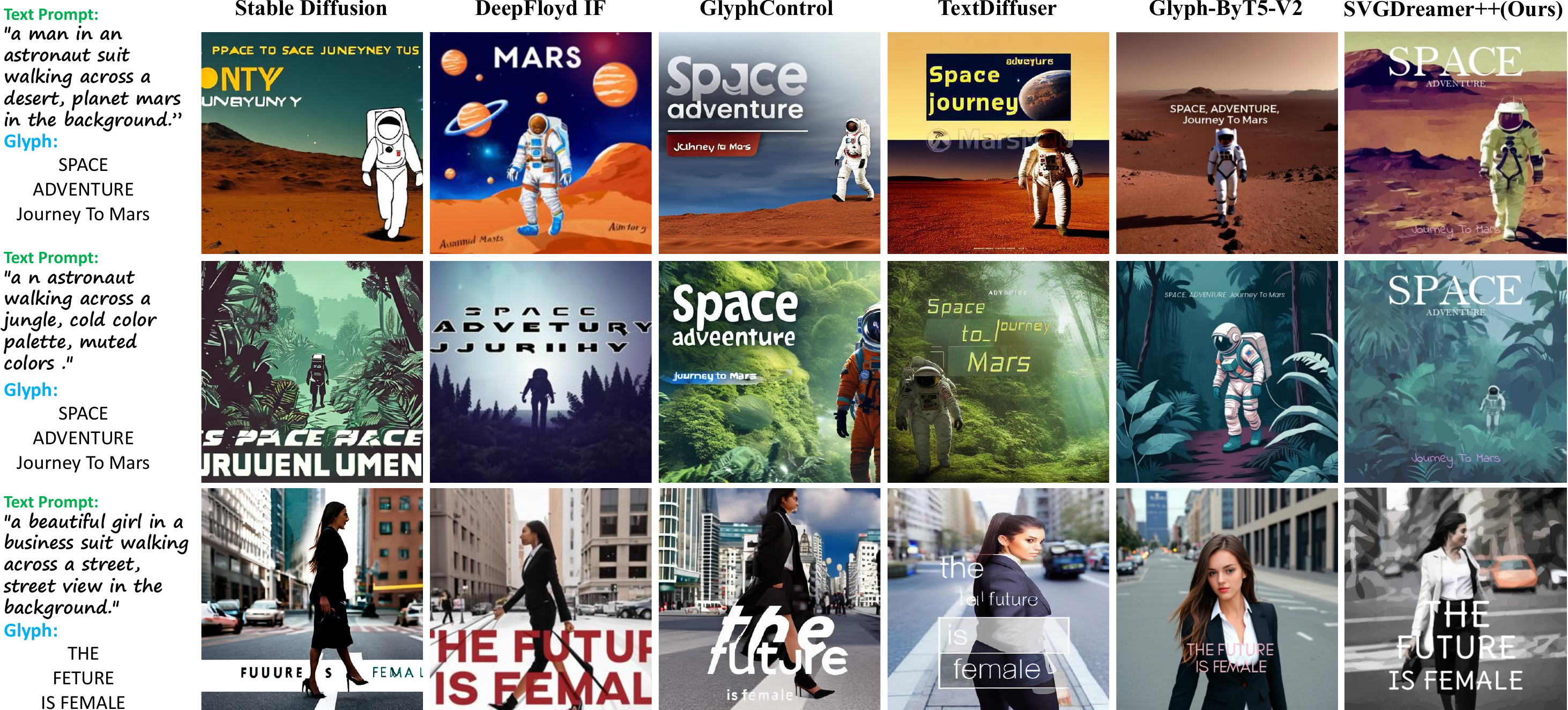}
\vspace{-2em}
\caption{
\textbf{Qualitative comparison between SVGDreamer++ vector poster synthesis and state-of-the-art raster poster synthesis methods}.
The column on the left represents the input text prompt used to generate the poster and the font symbols in the poster.
} \label{fig:poster_results}
\vspace{-1em}
\end{figure}
\begin{figure*}[t]
\centering
\includegraphics[width=0.9\linewidth]{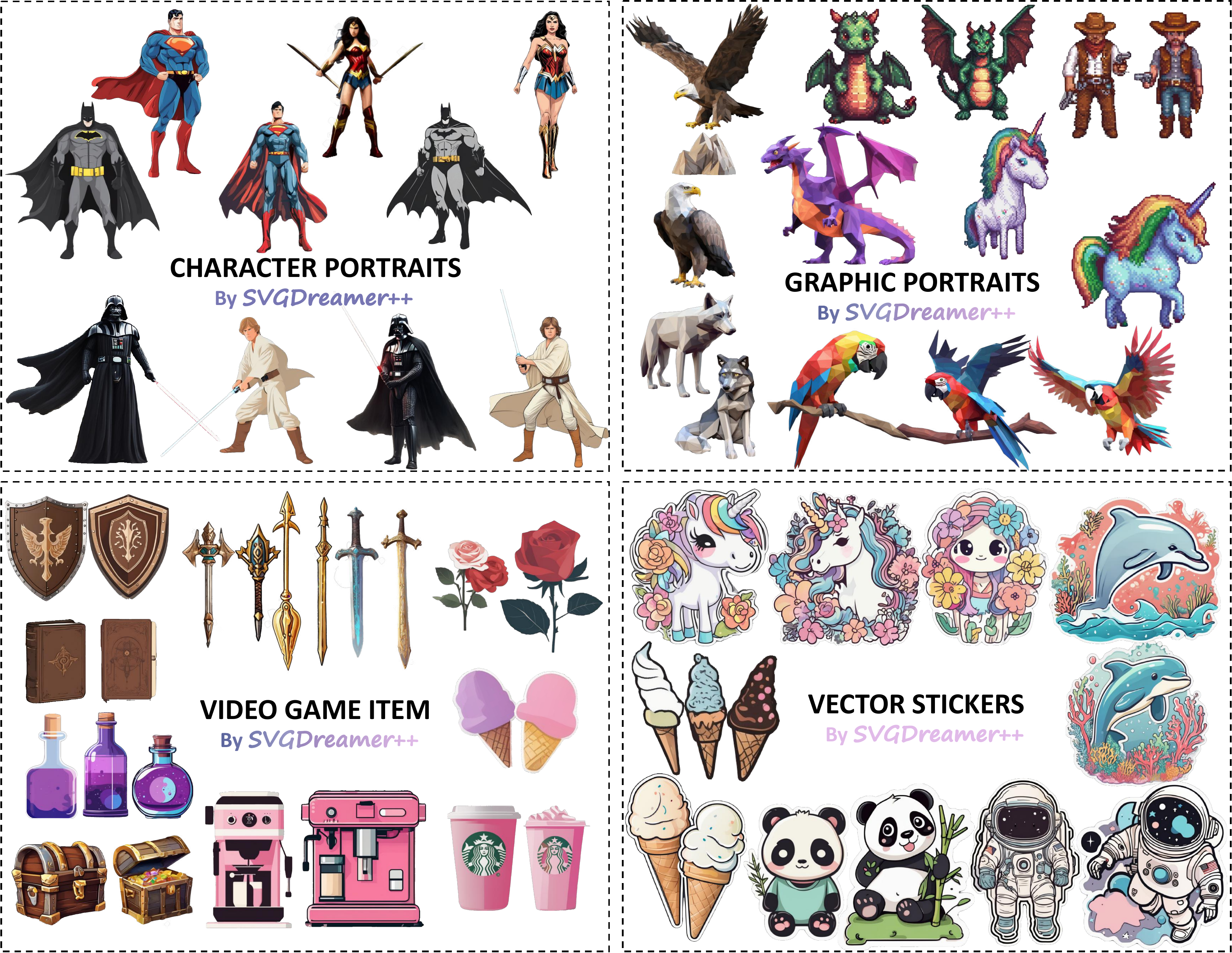}
\vspace{-1em}
\caption{
\textbf{The vector assets generated by SVGDreamer++}.
We present a curated collection of vector assets encompassing four distinct styles: character portraits, graphic portraits, video game items, and vector stickers.  Leveraging text descriptions, SVGDreamer++ can generate an extensive array of high-quality vector assets, which hold significant potential for application in the design industry.
}
\label{fig:vector_assets}
\vspace{-1em}
\end{figure*}
\subsection{Applications of SVGDreamer++}
\label{sec:application}
\subsubsection{Poster Design}
\label{sec:poster_desgin}
A poster is a large sheet used for advertising events, films, or conveying messages to people. It usually contains text and graphic elements. While existing T2I models have been developing rapidly, they still face challenges in text generation and control. On the other hand, SVG offers greater ease in text control. We will start by explaining the usage of our SVGDreamer++ tool for poster design. Initially, we employ SVGDreamer++ to generate graphic content. Then, we utilize modern font libraries to create vector fonts, taking advantage of SVG's transform properties to precisely control the font layout. Ultimately, we combine the vector images and fonts to produce comprehensive vector posters.
To be more specific, we employ the FreeType font library (\url{http://freetype.org/index.html}) to represent glyphs using vectorized graphic outlines. In simpler terms, these glyph's outlines are composed of lines, Bézier curves, or B-Spline curves. This approach allows us to adjust and render the letters at any size, similar to other vector illustrations.
The joint optimization of text and graphic content for enhanced visual quality is left for future work.

In Fig.~\ref{fig:poster_results}, we compare the posters generated by our SVGDreamer++ with those produced by five T2I models (All results generated by these T2I models are in raster format).
As depicted in Fig.~\ref{fig:poster_results}, both Stable Diffusion~\cite{ldm_Rombach_2022} (the 2nd column) and DeepFloyd IF~\cite{deepfloydif_stability_2023} (the 3rd column) display various text rendering errors, including missing glyphs, repeated or merged glyphs, and misshapen glyphs.
GlyphControl~\cite{glyphcontrol_yang_2023} (the 4th column) occasionally omits individual letters, and the fonts obscure content, resulting in areas where the fonts appear to lack content objects.
TextDiffuser~\cite{textdiffuser_chen_2023} (the 5th column) is capable of generating fonts for different layouts, but it also suffers from the artifact of layout control masks, which disrupts the overall harmony of the content. 
Glyph-ByT5-V2~\cite{glyphByT5v2_liu_2024} (the 6th column) enhances the aesthetic score of posters and is capable of controlling the overall layout of fonts within the specified bounding box. However, its control over the finer details of the font remains imprecise.
In contrast, posters created using our SVGDreamer++ are not restricted by resolution size, ensuring the text remains clear and legible. Moreover, our approach offers the convenience of easily editing both fonts and layouts, providing a more flexible poster design approach.

\subsubsection{Creative Vector Assets}
\label{sec:vector_assets}
The creation of vector assets is a time-intensive process for designers, and the acquisition of these assets is often costly due to intellectual property protections.
We investigate the application of SVGDreamer++ in generating vector assets across various styles.
The proposed SVGDreamer++ framework is capable of generating vector graphics at both the object level and part-level, offering exceptional editability.
Consequently, vector objects are extracted from the nouns identified in the text descriptions to compose vector graphic assets.
As illustrated in Fig.~\ref{fig:vector_assets}, all graphical elements in the four examples are generated using SVGDreamer++. 
We present a curated collection of vector assets encompassing four distinct styles: character portraits, graphic portraits, video game items, and vector stickers.
In contrast to diffusion-based~\cite{GLIDE_2022_nichol,ldm_Rombach_2022,DALLE2_2022_ramesh,imagen_2022_saharia,sdxl_podell_2024} raster objects, vector objects generated by our approach support localized editing, are not constrained by resolution, and feature a compact file representation.
The vector assets generated by SVGDreamer++ are characterized by their exceptional versatility and precision, making them particularly suitable for complex design tasks that require scalable and editable graphics.
These vector elements can be seamlessly integrated into design applications, such as web and advertising design, thereby enhancing the efficiency and creativity of the design process.



%% file: sec/5_conclusion.tex
\section{Conclusion \& Discussion}
\label{sec:conclusion}
\noindent In this work, we have introduced \textit{SVGDreamer++}, an innovative model for text-guided vector graphics synthesis. 
\textit{SVGDreamer++} improves on the previous state-of-the-art SVGDreamer in two ways.
Firstly, we introduce an advanced Hierarchical Image VEctorization algorithm, termed HIVE. This algorithm integrates an image segmentation prior to ensure more precise vectorization supervision, thereby rectifying the inaccurate boundaries observed in vector objects generated by SIVE.
Secondly, we propose a novel Adaptive Vector Primitives Control algorithm during the optimization phase to address and improve regions with deficient geometric features.
These empower our model to generate vector graphics with high editability, superior visual quality, and notable diversity. \textit{SVGDreamer++} is expected to significantly advance the application of text-to-SVG models in the design field.

\noindent\textbf{Limitations}. The editability of our method, which depends on the text-to-image (T2I) model used, is currently limited. However, future advancements in T2I diffusion models could enhance the decomposition capabilities of our approach, thereby extending its editability. Moreover, exploring ways to automatically determine the number of control points at the SIVE object level is valuable. 
